\documentclass[10pt,twocolumn,letterpaper]{article}

\usepackage{wacv}
\usepackage{times}
\usepackage{epsfig}
\usepackage{graphicx}
\usepackage{amsmath}
\usepackage{amssymb}

\usepackage[inline]{enumitem}
\usepackage{gensymb}

\usepackage{amsmath}
\usepackage{amsthm}
\usepackage{thmtools}
\usepackage{mathtools}
\usepackage{graphicx,subfigure}
\usepackage{verbatim}
\usepackage{multirow}
\usepackage{nicefrac}
\usepackage{url}
\usepackage{booktabs}
\usepackage{xspace} 

\usepackage{xcolor}
\usepackage{soul}
\newcommand{\acronym}[1]{\texterm{#1}\xspace}

\newcommand{\texterm}[1]{\textit{#1}}

\newcommand{\symbolmaxregularangle}{\alpha_R}

\newcommand{\symbollength}{l}
\newcommand{\symbolregularlength}{\symbollength_R}
\newcommand{\symboltopfilterlength}{\symbolregularlength^*}

\makeatletter
\def\instring#1#2{TT\fi\begingroup
  \edef\x{\endgroup\noexpand\in@{#1}{#2}}\x\ifin@}
\makeatother

\newcommand{\indexa}{i}
\newcommand{\indexb}{j}
\newcommand{\indexc}{k}

\newcommand{\projectpt}[2]{\tilde{{#1}_{#2}}}

\newcommand{\indexta}{a}

\newcommand{\amountb}{m}

\newcommand{\indexsymbolsevmpld}{v}
\newcommand{\indexsymbolplmatch}{s}

\newcommand{\initindex}{o}
\newcommand{\iindex}{\initindex}
\newcommand{\otherindexa}{\indexa}
\newcommand{\oindexa}{\otherindexa}
\newcommand{\otherindexb}{\indexb}
\newcommand{\oindexb}{\otherindexb}
\newcommand{\otherindexc}{\indexc}
\newcommand{\oindexc}{\otherindexc}

\newcommand{\initindext}{s}
\newcommand{\iindext}{\initindext}
\newcommand{\otherindexta}{\indexta}
\newcommand{\oindexta}{\otherindexta}

\newcommand{\customsymbol}[1]{#1}

\newcommand{\defset}[2]{#1~=~\{#2\}}

\newcommand{\setcardinality}[1]{\left\vert{#1}\right\vert}

\newcommand{\makepair}[2]{\left( #1, #2\right)}

\newcommand{\amountofviews}{N}

\newcommand{\symbolfocallength}{\customsymbol{f}}

\newcommand{\sfl}{\symbolfocallength}

\newcommand{\symbolstructurefrommotion}{\acronym{SfM}}
\newcommand{\ssfm}{\text{\textit{\symbolstructurefrommotion}}~}

\newcommand{\pointgen}{x}

\newcommand{\point}[1]{\customsymbol{\pointgen_{#1}}}

\newcommand{\ipoint}{\point{\iindex}}

\newcommand{\imga}{\img{\indexa}}
\newcommand{\imgb}{\img{\indexb}}

\newcommand{\iimg}{\img{\iindex}}
\newcommand{\oimga}{\img{\oindexa}}
\newcommand{\oimgb}{\img{\oindexb}}
\newcommand{\oimgc}{\img{\oindexc}}

\newcommand{\symbolcamera}{C}
\newcommand{\cam}{\symbolcamera}
\newcommand{\symbolcamerapos}{c}
\newcommand{\campos}{\symbolcamerapos}

\newcommand{\icam}{\cam_\iindex}

\newcommand{\icampos}{\campos_\iindex}
\newcommand{\ocamposa}{\campos_\oindexa}

\newcommand{\img}[1]{\customsymbol{I_{#1}}}

\newcommand{\symbolline}{l}

\newcommand{\symbolepipolarline}{\symbolline}
\newcommand{\sepl}{\symbolepipolarline}
\newcommand{\sepla}{\sepl_\indexa}

\newcommand{\iepl}{\sepl_\iindex}

\newcommand{\feq}[1]{
  \begin{equation}
    #1
  \end{equation}
}

\newcommand{\symbolradius}{r}

\newcommand{\sectionname}{Section}
\newcommand{\sref}[1]{\sectionname~\ref{#1}}

\newcommand{\cref}[1]{\chaptername~\ref{chapter#1}}

\newcommand{\fref}[1]{\figurename~\ref{#1}}

\newcommand{\symboldistance}{d}

\newcommand{\symbolpolyline}{\gamma}

\newcommand{\spcomma}{\text{~},}

\newcommand{\spl}{\symbolpolyline}

\newcommand{\spla}{\symbolpolyline_\indexa}
\newcommand{\splb}{\symbolpolyline_\indexb}

\newcommand{\funcdistance}[2]{\left\vert \left\vert #1 - #2 \right\vert \right\vert}
\newcommand{\fdist}[2]{\funcdistance{#1}{#2}}

\newcommand{\mediannumberobservations}{v_M}

\newcommand{\symbolpotentialcorrespondence}{h}
\newcommand{\spc}{\symbolpotentialcorrespondence}
\newcommand{\symbolpotentialcorrespondencesset}{H}
\newcommand{\spcs}{\symbolpotentialcorrespondencesset}
\newcommand{\symbolpotentialcorrespondencessetsset}{\spcs}
\newcommand{\spcss}{\symbolpotentialcorrespondencessetsset}

\newcommand{\sipcs}{\spcs_\iindex}
\newcommand{\sopcsa}{\spcs_\oindexa}
\newcommand{\sopcsb}{\spcs_\oindexb}

\newcommand{\sopcan}[1]{\spc_{\oindexa#1}}

\newcommand{\sopccn}[1]{\spc_{\oindexc#1}}

\newcommand{\sopca}{\spc_{\oindexa\indexta}}

\newcommand{\sospcsa}{\spcs_\indexa}

\newcommand{\symbolreferencepoint}{p}
\newcommand{\rpt}{\symbolreferencepoint}

\newcommand{\irpt}{\rpt_\iindex}
\newcommand{\orpta}{\rpt_\oindexa}

\newcommand{\symbolsphere}{\mathcal{S}}
\newcommand{\symbolinner}{I}
\newcommand{\symbolouter}{O}

\newcommand{\symbolsearchspereouter}{\symbolsphere_\symbolouter}
\newcommand{\spho}{\symbolsearchspereouter}
\newcommand{\symbolsearchsphereradiusouter}{\symbolradius_\symbolouter}

\newcommand{\rado}{\symbolsearchsphereradiusouter}

\newcommand{\symbolsearchspereinner}{\symbolsphere_\symbolinner}
\newcommand{\sphi}{\symbolsearchspereinner}
\newcommand{\symbolsearchsphereradiusinner}{\symbolradius_\symbolinner}

\newcommand{\radi}{\symbolsearchsphereradiusinner}

\newcommand{\symbolcircle}{\mathcal{O}}
\newcommand{\cir}{\symbolcircle}

\newcommand{\icir}{\cir_\iindex}
\newcommand{\ocira}{\cir_\oindexa}

\newcommand{\irad}{\symbolradius_\iindex}
\newcommand{\orada}{\symbolradius_\oindexa}

\newcommand{\ssep}{\eponi}

\newcommand{\sisep}{\ssep} 

\newcommand{\amountcorrespondingeps}{\amountb}

\newcommand{\symbolpolylinematchesset}{M}
\newcommand{\spms}{\symbolpolylinematchesset}
\newcommand{\symbolpolylinematchessetsset}{\spms}
\newcommand{\spmss}{\symbolpolylinematchessetsset}

\newcommand{\spmsn}[1]{\spms_{#1}}

\newcommand{\sopmsa}{\spms_\oindexa}

\newcommand{\edgepoint}{x}
\newcommand{\ep}{\edgepoint}
\newcommand{\eptd}{\ep} 

\newcommand{\eponi}{\ep_\iindex}
\newcommand{\epona}{\ep_\oindexa}

\newcommand{\iep}{\edgepoint^\iindext}
\newcommand{\oepa}{\edgepoint^\oindexta}

\newcommand{\iiep}{\iep_\iindex}
\newcommand{\ioepa}{\iep_\oindexa}

\newcommand{\oaiep}{\oepa_\iindex} 
\newcommand{\oaoepa}{\oepa_\oindexa} 
\newcommand{\oaoepb}{\oepa_\oindexb}

\newcommand{\matchdirectionfollowlength}{\symbollength_d}

\newcommand{\xoverbrace}[2][\vphantom{\dfrac{A}{A}}]{\overbrace{\vphantom{#1}#2}}
\newcommand{\xunderbrace}[2][\vphantom{\dfrac{A}{A}}]{\underbrace{\vphantom{#1}#2}}

\newcommand{\spltd}{\spl}

\newcommand{\expandvisibilitymaxpldist}{\symboldistance_\indexsymbolsevmpld}

\newcommand{\symbolpolylinematchingrefpointpldist}{\symboldistance_\indexsymbolplmatch}

\newcommand{\symbolcloserefpoints}{P}
\newcommand{\closerefpoints}[1]{\symbolcloserefpoints^{#1}}

\newcommand{\symbolweight}{w}
\newcommand{\refpointweight}[1]{\symbolweight_{#1}}

\newcommand{\mininlierobservations}{k_v}

\newcommand{\pepc}{\acronym{PEPC}}
\newcommand{\pepcs}{\acronym{PEPCs}}

\newcommand{\pec}{\acronym{PEC}}
\newcommand{\pecs}{\acronym{PECs}}

\declaretheoremstyle[%
  spaceabove=6pt,%
  spacebelow=6pt,%
  headfont=\normalfont\bfseries,%
  postheadspace=0.5em,%
  qed={},%
  headpunct={}
]{mystyle} 
\declaretheorem[name={Definition},style=mystyle,unnumbered,
]{Definition}

\newcommand{\simpol}[2]{\text{s}\left(#1, #2\right)}
\newcommand{\simpolstd}{\simpol{\spla}{\splb}}

\newcommand{\sysname}{EdgeGraph3D\xspace}

\newcommand{\repolink}{https://github.com/abignoli/EdgeGraph3D}

\wacvfinalcopy 

\def\wacvPaperID{****} 

\ifwacvfinal\fi
\begin{document}

\title{Multi-View Stereo 3D Edge Reconstruction}

\author{Andrea Bignoli \hspace{0.3cm} Andrea Romanoni \hspace{0.3cm} Matteo Matteucci \\
	Politecnico di Milano\\
	{\tt\small andrea.bignoli@mail.polimi.it \hspace{0.2cm} andrea.romanoni@polimi.it \hspace{0.2cm} matteo.matteucci@polimi.it}
}

\maketitle
\ifwacvfinal\thispagestyle{empty}\fi

\begin{abstract}
This paper presents a novel method for the reconstruction of 3D edges in multi-view stereo scenarios. Previous research in the field typically relied on video sequences and limited the reconstruction process to either straight line-segments, or edge-points, \ie, 3D points that correspond to image edges. We instead propose a system, denoted as \sysname, able to recover both straight and curved 3D edges from an unordered image sequence. A second contribution of this work is a graph-based representation for 2D edges that allows the identification of the most structurally significant edges detected in an image. We integrate \sysname in a multi-view stereo reconstruction pipeline and analyze the benefits provided by 3D edges to the accuracy of the recovered surfaces. We evaluate the effectiveness of our approach on multiple datasets from two different collections in the multi-view stereo literature. Experimental results demonstrate the ability of \sysname to work in presence of strong illumination changes and reflections, which are usually detrimental to the effectiveness of classical photometric reconstruction systems.
\end{abstract}

\section{Introduction}
\label{sec:intro}
Reconstructing the 3D shape of a scene captured by a set of images represents a long-standing problem faced by the computer vision community. 
Structure from Motion methods address the simultaneous estimation of camera positions and orientations together with a point-based representation of the environment \cite{wu13,mo13,mouragnon_et_al07,snavely_et_al06}.
Multi-View Stereo algorithms usually bootstrap from such estimations to recover a mesh-based dense reconstruction.

State-of-the-art mesh-based algorithms \cite{vu_et_al_2012,li2015detail,li2016efficient,romanoni2017multi} are initialized through Delaunay-based space carving algorithms such as \cite{tola12,litvinov_lhuillier_13,hoppe13,romanoni15b} which estimate a mesh  from the  structure from motion points or from dense point clouds computed through depth maps. 
The authors of \cite{romanoni15b} showed that the Delaunay triangulation built upon 3D edge-points, \ie, points belonging to 3D edges, is able to  represent the shape of the environment better than using 3D points reconstructed from classical 2D features. 
The usage of 3D edges or 3D edge-points presents two significant benefits: they are robust to significant illumination changes that can negatively affect standard photometric-based depth maps estimation, and they are a compact representation of the salient part of a scene, i.e., they avoid redundancies along flat surfaces.

The reconstruction of 3D edges can be performed by matching directly their 2D observations across a sequence of images.
This is a challenging task since corresponding edges often cannot be matched just on the basis of their geometric parameters.
In literature, edge reconstruction is often limited to the reconstruction of line-segments, \ie, straight edges, and existing approaches rely their estimation on video sequences.
Only 
Hofer \etal \cite{hofer2015line3d}  with Line3D++ proposed an approach to estimate 3D segments in a Multi-View Stereo scenario. 
This method, however, is not able to recover curved edges.

In this paper we propose a novel algorithm for the reconstruction of 3D edges, both straight and curved, from an unordered set of images. Furthermore, we illustrate how the points belonging to 3D edges can significantly improve the appearance and the accuracy of the 3D models reconstructed from sparse point clouds, using the algorithm proposed by \cite{litvinov_lhuillier_13} and improved by \cite{romanoni15b}.
We tested our algorithms on the fountain-p11 dataset provided in the EPFL Multi-View Stereo collection \cite{strecha2008} and on the recent DTU dataset
\cite{jensen2014large}.

In Section \ref{sec:related} we overview the state-of-the art of 3D edge reconstruction.
In Section \ref{subsec:2dedge} we show the 2D edge representation we use in our algorithm.
In Section \ref{sec:algo} we describe the proposed method to reconstruct 3D edges and introduce the \sysname system.
In Section \ref{sec:exp} we discuss the results of our approach on two well-known datasets.
In Section \ref{sec:concl} we conclude the paper and we illustrate some possible future research directions.

\section{Related works}
\label{sec:related}

\begin{figure*}[t]
     \centering
     \hfill
    \begin{minipage}{0.47\textwidth}
        \centering
        \textbf{before filtering}
        \includegraphics[width=\textwidth]{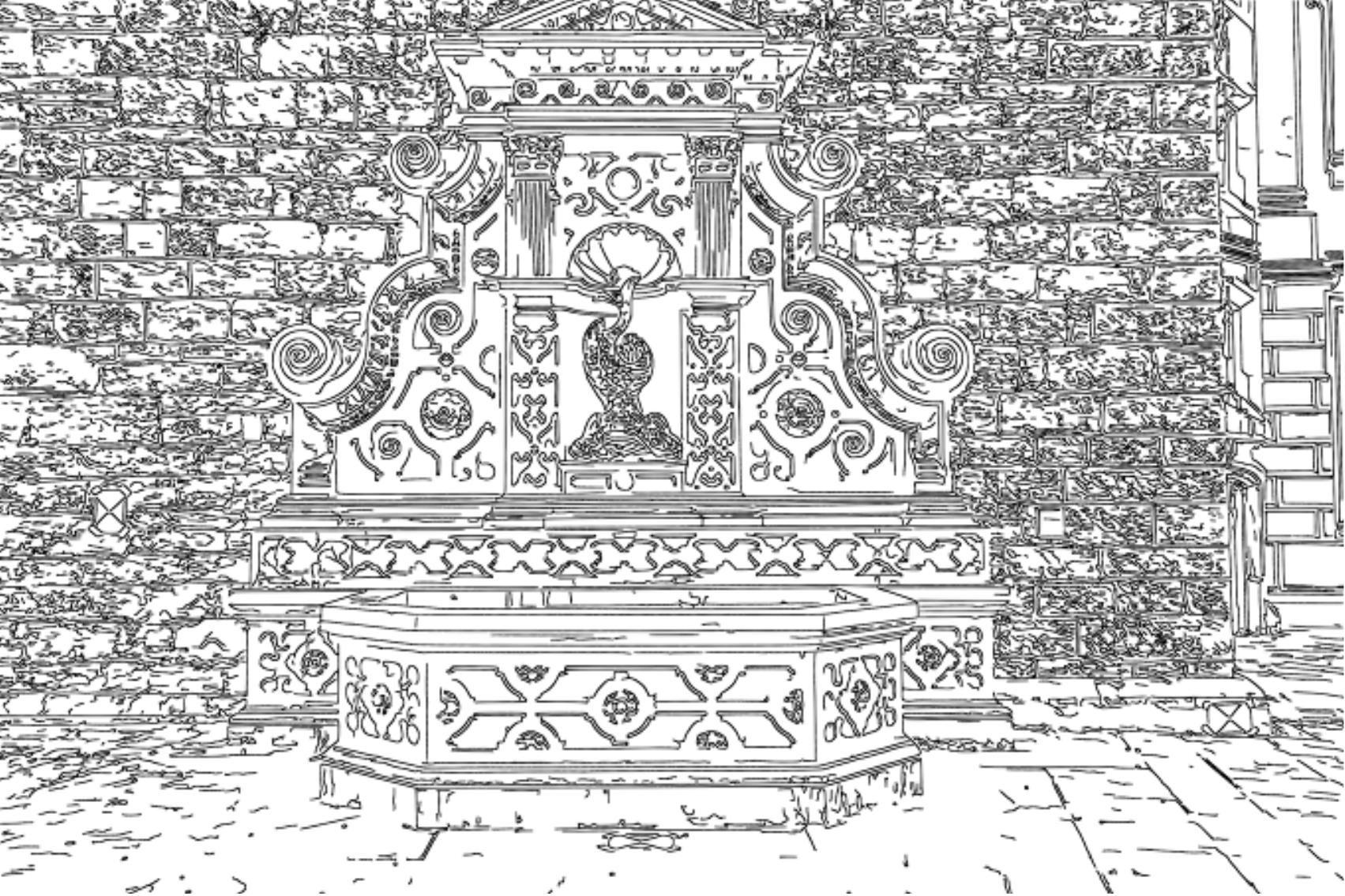} 
    \end{minipage}\hfill
    \begin{minipage}{0.47\textwidth}
        \centering
        \textbf{after filtering}
        \includegraphics[width=\textwidth]{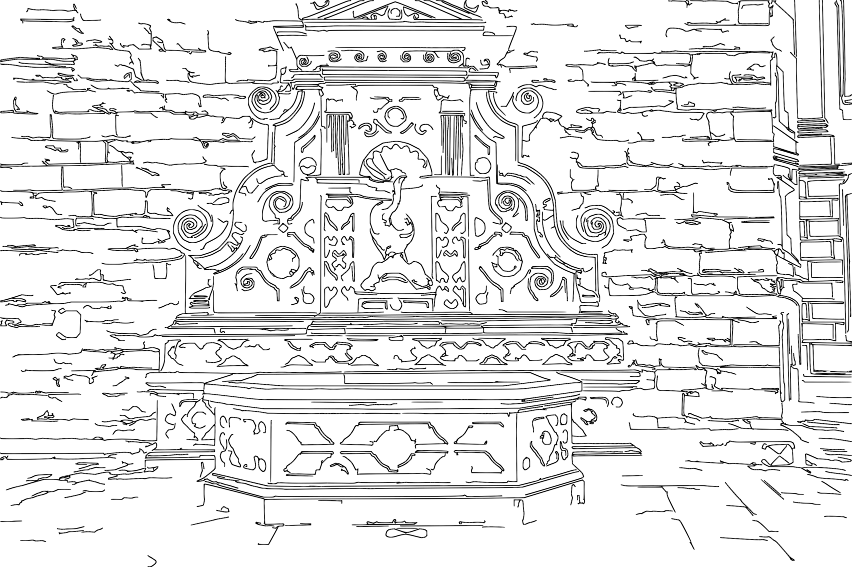} 
    \end{minipage}\hfill
    \hfill
    \vspace{\belowdisplayskip}
    \caption{Edge-graph filtering: (left)~original edge-graph, generated from an edge-image produced by the algorithm presented in \cite{meer2001} (right)~filtered edge-graph, in which non-structural edges have been removed}
	\label{fig:filtering}
\end{figure*}
	
In literature, a limited amount of works address directly the issue  of edge reconstruction, and usually they recover only long straight edges. 
Tian \etal \cite{tian2008automatic} track 2D points belonging to images edges along a video sequence and estimate their 3D position, then they recover the 3D edges that connect these 3D points.
Zhang and Baltsavias \cite{zhang2000knowledge} propose an algorithm for the reconstruction of road edges from aerial images. They match only straight edges, since these are prevalent in urban environments, by using epipolar geometry.

Edges have been also integrated in structure from motion pipelines as elements to robustly recover or optimize the camera poses.
In \cite{taylor1995structure}, the authors propose a structure from motion algorithm based on-line segments correspondences in an image sequence. 
Instead of classical point-based reprojection error measure, they optimize a non-linear objective function that measures the total squared distance between the observed edge segments and the projections of the reconstructed lines on the image plane. 
Ansar \etal \cite{ansar2003linear} propose the mathematical foundations of a camera pose estimation systems able to use both points and lines for real-time camera pose estimation using linear
optimization, while \cite{eade2009edge} proposes a monocular Simultaneous Localization and Mapping (SLAM) algorithm which adopts as landmarks small edges, \ie, the edgelets,  instead of the classical point-based features.
Also the latter algorithms are able to estimate 3D straight edges, while curved edges are neglected, moreover they rely on video sequences.

In \cite{romanoni15b}, the authors estimate the 3D position of 2D points belonging to images edges,  and they embedded them into a Delaunay triangulation to make the triangulation edges closer to the real 3D structure of the scene.
In this case even points on curved edges are taken into account, but the actual 3D edges are not explicitly reconstructed.
Similarly, Bodis \etal \cite{bodis2017efficient} present a reconstruction algorithm for large-scale multi-view stereo, able to produce meshes that are consistent with the bidimensional edges of the input images; they enforce the Delaunay Triangulation, employed in the reconstruction, to be properly
divided along edges.
While the purpose of the system is not the actual reconstruction of the 3D edges, they further show the benefits that edges offer to the reconstruction process by sharply defining the architectural elements in urban scenes. 

This review shows that edges or the points belonging to edges are sometimes adopted to improve the robustness of 3D reconstruction, camera tracking or SLAM algorithms.
However these techniques deal only with video sequences to simplify the process of edge matching and they focus only on straight edges estimation; a general approach to 3D edge reconstruction from unordered set of images is thus the novel contribution of this paper that is also beneficial for more complex dense 3D reconstruction algorithms.

\section{2D Edge-Graphs}
\label{subsec:2dedge}

Decades of research in computer vision produced several edge detection algorithms.
Many of the proposed techniques, such as \cite{canny_86}, are able to describe both straight and curved edges through an image output, but they represent the edges only at pixel level. On the other hand, other approaches as \cite{gioi2012}, estimate line-segments, which describe edges with subpixel accuracy, but are not able to properly represent and detect curved edges. 
Here, we introduce an alternative representation for 2D edges, able to both represent curved edges and to reach sub-pixel accuracy.
We propose to use an undirected graph, named \texterm{2D edge-graph}, in which nodes represent 2D points on an image, and connections between nodes indicate detected edges connecting their extremes. 
An additional benefit provided by the use of edge-graphs is the direct description of the connections between different edges detected in an image, which will be key to the techniques presented in \sref{sec:pepcgen}.

  \begin{figure*}[t]
    \begin{center}
      \includegraphics[width=\textwidth]{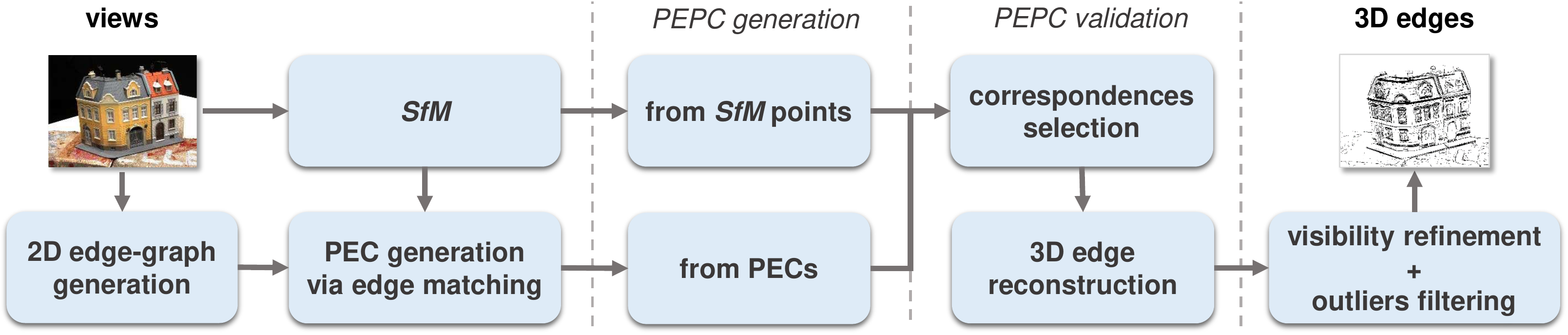}
    \end{center}
    \caption{Proposed 3D edge reconstruction pipeline}
    \label{fig:proposed-pipeline}
  \end{figure*}

We  generate an edge-graph from the edge-images produced by the standard edge-detection algorithms presented in \cite{meer2001}. We assign a node to the center of each edge-pixel; then, for each pair of adjacent edge-pixels, we connect the corresponding nodes if it does not generate small, undesired, loops of length shorter than 4 \emph{px}, \eg, which do not represent meaningful connections between edges in the original edge-image. 
Since edges recovered using this process still follow the discretized structure described by the original edge-image, we apply a \texterm{polyline smoothing} step.
A polyline is a sequence of edges in the graph, in which all intermediate nodes have exactly two connections. Polyline smoothing is a process by which the original polyline is transformed so that: 
\begin{enumerate*}[label={\roman*)},font={\color{black}\bfseries}]
\item the extremes of the polyline are left unaltered
\item the sequence of intermediate nodes is modified to the shortest sequence guaranteeing that a distance no greater than 1 pixel separates the original polyline from the final one
\end{enumerate*}. We employed a variant of the Douglas-Peucker~\cite{douglas1973} algorithm to achieve this result. Using this technique, it is possible to obtain a suitable representation for edges in a scene, at subpixel-accuracy.

Many standard edge detection algorithms generate a significant number of edges that do not correspond to structural elements in the scene. 
To filter them out, we propose an edge-graph filtering step which retains only long polylines composed by line-segments without significant sharp variations in direction. 
Let us define:
\begin{Definition}
\textbf{Regular length} 
The \textit{regular length} of a polyline is the length of the longest interval of connected line-segments for which each angle between consecutive elements is not greater than a fixed threshold $\symbolmaxregularangle$ (\eg,  $\symbolmaxregularangle~\approx~20\degree$).
\end{Definition}

For each 2D edge-graph we rank all its polylines according to their regular length and 
we compute  $\symboltopfilterlength$ as the shortest length among the top 10\% of these polylines. 
We then filter out all the connected components in the graph that do not contain a polyline with regular length greater than $\symboltopfilterlength$. In \fref{fig:filtering}, we show that structural edges are preserved, while irrelevant edges are almost completely filtered out.

\section{The \sysname system}
\label{sec:algo}
\label{subsec:overview}

\sysname is able to reconstruct three-dimensional edges from their observations in the input views. Curved edges are represented as 3D polylines, \ie, a sequence of straight 3D line-segments connected to each other.
The inputs of our method are the 2D edge graphs computed for each image, the camera calibrations and an initial set of 3D points estimated through structure from motion.
To understand the key idea behind the proposed algorithm, let us assume a 3D edge-point and its 2D observations on a subset of images are provided; these observations likely lie on 2D edges. 
We simultaneously follow such 2D edges on all the images involved, generating a sequence of corresponding 2D edge-points that identify new 3D edge-points. The sequence of the recovered 3D edge-points defines a 3D polyline representing the reconstructed 3D edge.

 \begin{figure*}[t]
     \centering
     \begin{minipage}{0.45\textwidth}
     \begin{minipage}{0.45\textwidth}
     \includegraphics[width=\textwidth]{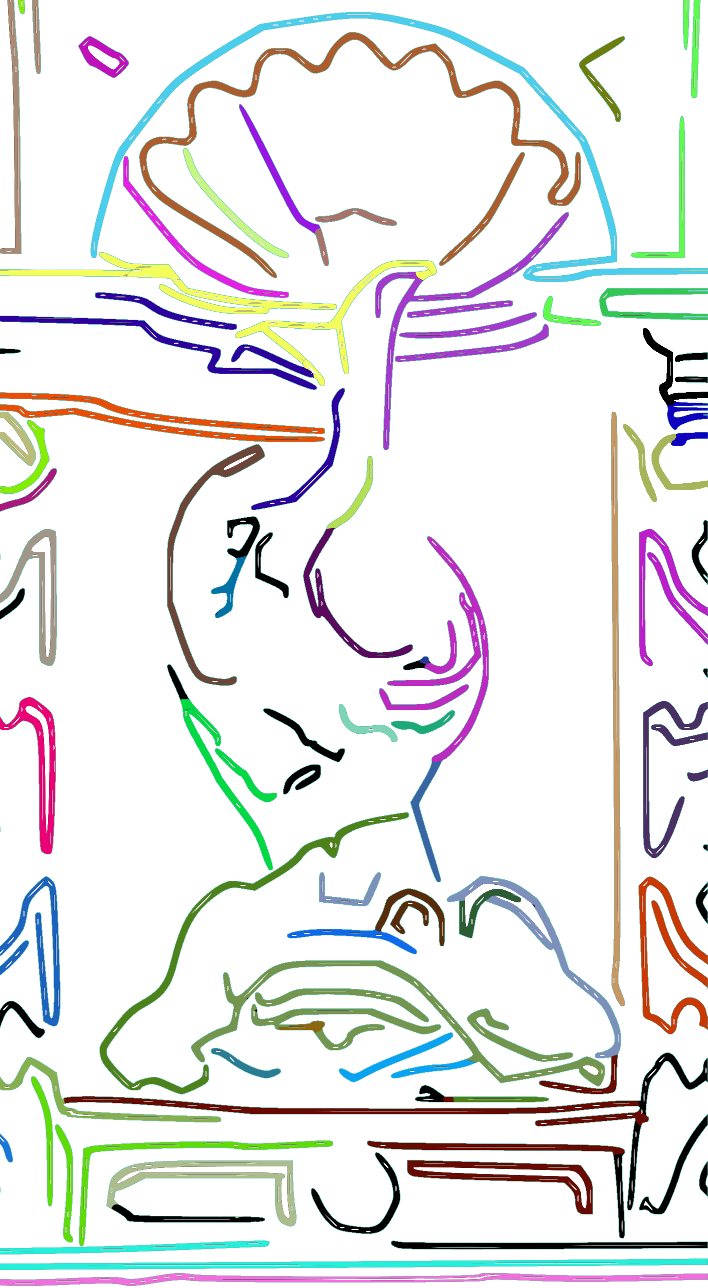}
     \end{minipage}\hfill
     \begin{minipage}{0.45\textwidth}
     \includegraphics[width=\textwidth]{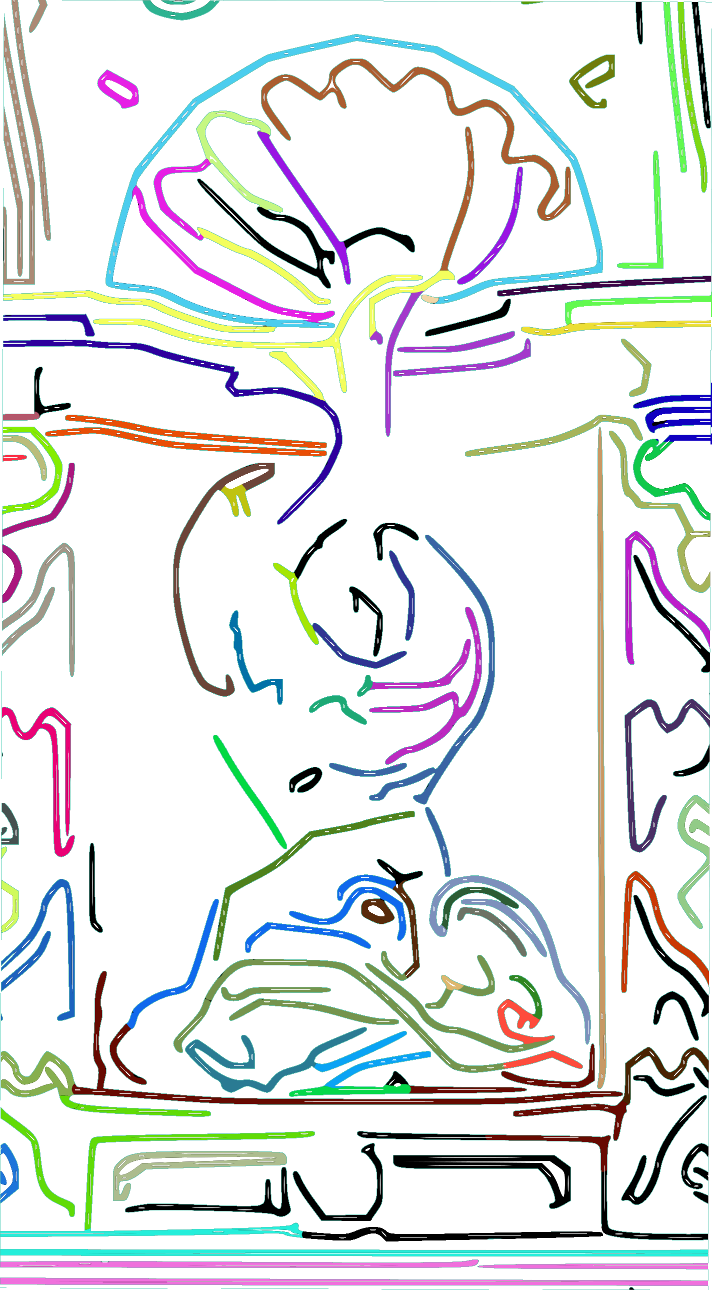}
     \end{minipage}
     \label{fig:edge-matching}
     \vspace{\belowdisplayskip}
     \caption{Visualization of the output of the edge matching procedure on two 2D edge-graphs of the fountain-P11~\cite{strecha2008} dataset. Polylines with the same color belong to the same \pec. This sample visually show the effectiveness of the approach, regardless of the incompleteness of the detected edges on both images.}
     \end{minipage}\hfill
     \begin{minipage}{0.45\textwidth}
     \centering
     \begin{minipage}{\textwidth}
    \subfigure[]{
\begin{minipage}{\textwidth}
    \begin{minipage}{0.32\textwidth}
        \centering
        \textbf{image $I_0$}
        \includegraphics[width=\textwidth]{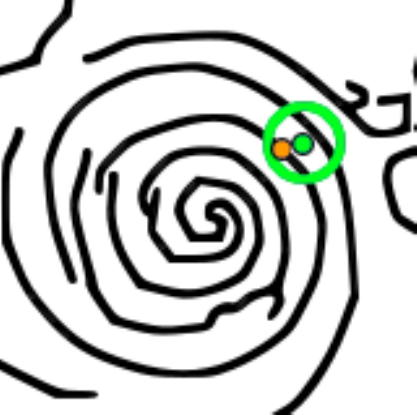}\\
    \end{minipage}\hfill
    \begin{minipage}{0.32\textwidth}
        \centering
        \textbf{image $I_1$}
        \includegraphics[width=\textwidth]{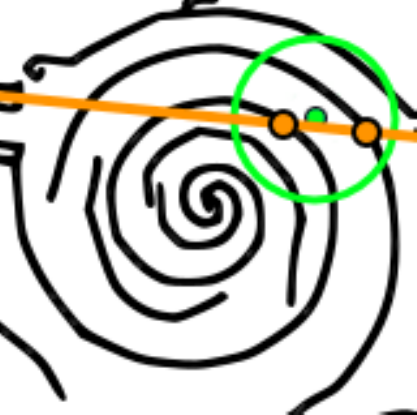}\\
    \end{minipage}\hfill
    \begin{minipage}{0.32\textwidth}
        \centering
        \textbf{image $I_2$}
        \includegraphics[width=\textwidth]{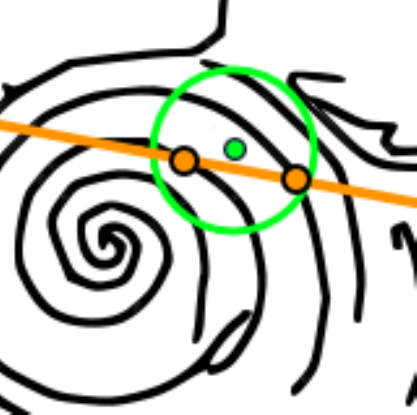}\\
    \end{minipage}
\end{minipage}
    \vspace{\belowdisplayskip}
    \label{fig:1a}
    }
    \end{minipage}
    \begin{minipage}{\textwidth}
    \subfigure[]{
    \begin{minipage}{\textwidth}
    \begin{minipage}{0.32\textwidth}
        \centering
        \textbf{image $I_0$}
        \includegraphics[width=\textwidth]{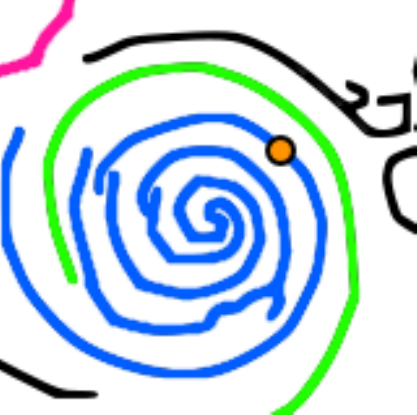}\\
    \end{minipage}\hfill
    \begin{minipage}{0.32\textwidth}
        \centering
        \textbf{image $I_1$}
        \includegraphics[width=\textwidth]{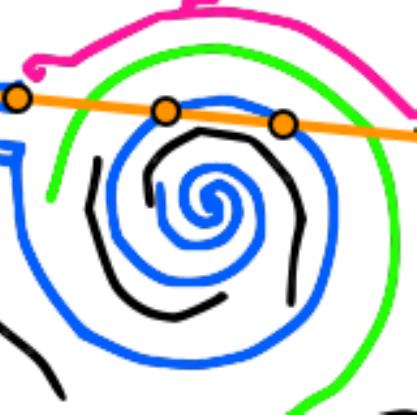}\\
    \end{minipage}\hfill
    \begin{minipage}{0.32\textwidth}
        \centering
        \textbf{image $I_2$}
        \includegraphics[width=\textwidth]{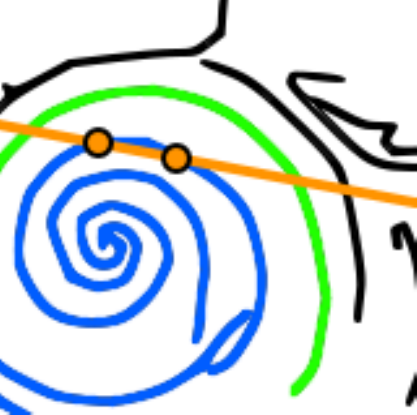}\\
    \end{minipage}
\end{minipage}
    \label{fig:1b}
    \vspace{\belowdisplayskip}
    }\hfill
    \end{minipage}
    
    \caption{Generation of \pepcs: \protect\subref{fig:1a} searching a new edge-point (in orange) in the vicinity of a known 3D point (in green) \protect\subref{fig:1b} searching a new edge-point using polyline matches, represented by edges shown in the same color on different images. A new orange 2D edge-point on the blue match is selected as target.}
    \end{minipage}
\end{figure*}

\subsection{System Overview}

In \fref{fig:proposed-pipeline} we illustrate the full pipeline of \sysname.
From the input images we compute the camera calibration via SfM.
For each image we define the corresponding 2D Edge-Graph presented in the previous section that we use to match edges on multiple views and to define
\begin{Definition}
\textbf{Potential edge correspondence (PEC)}
A \pec is a set of 2D polylines (\ie, image edges), on multiple views, that are considered projections the same 3D edges.
\end{Definition}

In \sref{sec:pepcgen} we illustrate how, for each image 2D edge-point, we exploit the epipolar constraint to generate 
\begin{Definition}
\textbf{Potential edge-point correspondence (PEPC)}
A \pepc is a set of possibly corresponding 2D edge-points on multiple views, which may even contain multiple points on the same view, from which it may be possible to generate a new 3D edge-point.
\end{Definition}
We bound the number of \pepc, by relying on the spatial information carried by the \ssfm 3D points and on the edge matches collected in the \pec s.
In \sref{sec:pepcval}, we validate each \pepc, while reconstructing 3D edges. 
Finally in \sref{sec:visibility} we illustrate how we improve the visibility information associated to a 3D edge and how we remove outliers. 

\subsection{Edge Matching and \pec generation}
\label{sec:edge-matching}

To compute \pecs let us consider a pair of 2D polylines $\spla$ and $\splb$ on images $\imga$ and $\imgb$; if a pair of polylines shares a significant amount of nearby 3D points, they must occupy nearby locations in the 3D space, and are potentially associated to the same 3D edges.
Therefore we define a similarity measure $\simpolstd$ for $\spla$ and $\splb$ as: 
\begin{equation}
\simpolstd = \frac{\sum\nolimits_{\substack{\rpt \in \closerefpoints{\spla} \cap \closerefpoints{\splb}}} \refpointweight{\rpt}}{\sum\nolimits_{\substack{\rpt \in \closerefpoints{\spla} \cup \closerefpoints{\splb}}} \refpointweight{\rpt}},
\end{equation}
where $\closerefpoints{\spla}$ is the list of 3D points visible on $\imga$, that lie within a distance of $\symbolpolylinematchingrefpointpldist$ from $\spla$, and the weight $\refpointweight{\rpt}$ of a point $\rpt$ is defined as the inverse of the average number of polylines close to the  reprojections of $\rpt$, where $\rpt$ is visible.

Then, we build a \textit{polyline similarity graph} as an undirected weighted graph, in which nodes represent different polylines on different images, and the weight of each edge is equal to the similarity of the polylines associated to its extremes. 
We then use the community detection algorithm in \cite{blondel2008fast}, on the  polyline similarity graph. Communities are subsets of nodes of a graph that are densely interconnected, hence in our case they are subsets of polylines, on multiple views, with high degree of similarity, \ie, they represent \pecs, as \fref{fig:edge-matching} shows.

\subsection{\pepc generation}
\label{sec:pepcgen}

To present the principle that inspires the two strategies for \pepcs generation presented in this section, let us consider a 2D edge-point $\ssep$ in the 2D edge-graph of image $\iimg$. 
To recover the corresponding 3D edge-point $\eptd$ we identify the potential 2D edge-points correspondences on other views through epipolar geometry. 
The correspondence $\epona$ on a second image $\oimga \neq \iimg$ must lie on the epipolar line $\sepla$ generated on $\oimga$ by $\ssep$.
Since we assume that the correspondence $\epona$ is a 2D edge-point, we generate a finite set of potential correspondences $\sopcsa~=~\{\sopcan1, \sopcan2, \dots \sopcan{\amountcorrespondingeps}\}$ by intersecting the 2D edge-graph associated to $\oimga$ with the epipolar line $\sepla$.
We repeat this process for all other views where the new potential 3D edge-point may be visible.
The cardinality of the sets of potential correspondences $\sopcsa$ is however generally too high to search for the correct correspondences on all views in acceptable computation time. Therefore, we propose two approaches to limit the set of potential correspondences on each view.

\subsubsection{From \ssfm points}

In the first approach (\fref{fig:1a}) we exploit the knowledge of a 3D point position $\rpt$, recovered through structure from motion, and we constrain the search for a new 3D edge-point $\eptd$ in its neighborhood, in particular, to a sphere $\spho$ centered in $\rpt$ with radius $\rado$.
Given an image $\iimg$ where $\rpt$ projects on a location $\irpt$, we search the initial 2D edge-point $\ipoint$ within the projection of a  sphere $\sphi$ centered in $\rpt$ with a radius $\radi < \rado$. This projection is an elliptic conic section that we approximate with a circle $\icir$ centered in $\irpt$ (green circle in image $\iimg$ of \fref{fig:1a}), of radius $\irad = \radi \frac{\fdist{\icampos}{\rpt}}{\sfl}\spcomma$
where $\icampos$ represents the center of camera $\icam$ that produced $\iimg$, and $\sfl$ is the focal length. 
Then, for each polyline passing through $\icir$, we select the 2D edge-point closest to $\rpt$ (orange point in image $\iimg$ of \fref{fig:1a}) as the initial edge-points for which we aim to find 2D to 2D correspondences on other views. 

For each image $\oimga \neq \iimg$ where $\rpt$ projects, we look for correspondences near the projection $\orpta$ of $\rpt$ on $\oimga$ (orange point in images $I_1$ and $I_2$ of \fref{fig:1a}). 
In particular, we constrain the search of correspondences within circle $\ocira$, centered in $\orpta$, of radius $\orada = \rado \frac{\fdist{\ocamposa}{\rpt}}{\sfl}\spcomma$
which approximates the projection of sphere $\spho$ on $\oimga$ (green circles in images $I_1$ and $I_2$ of \fref{fig:1a}). 
Since the correspondences of $\ssep$ on $\oimga$ are bound to lie on the epipolar line $\iepl$, we combine both constraint looking for intersections between $\iepl$ and the edge-graph in $\ocira$, to determine the set of potential correspondences $\sopcsa$ of $\ssep$ on image $\oimga$. Repeating the same process on multiple views generates a new \pepc.

  \begin{figure*}[t]
    \begin{center}
      \includegraphics[width=\textwidth]{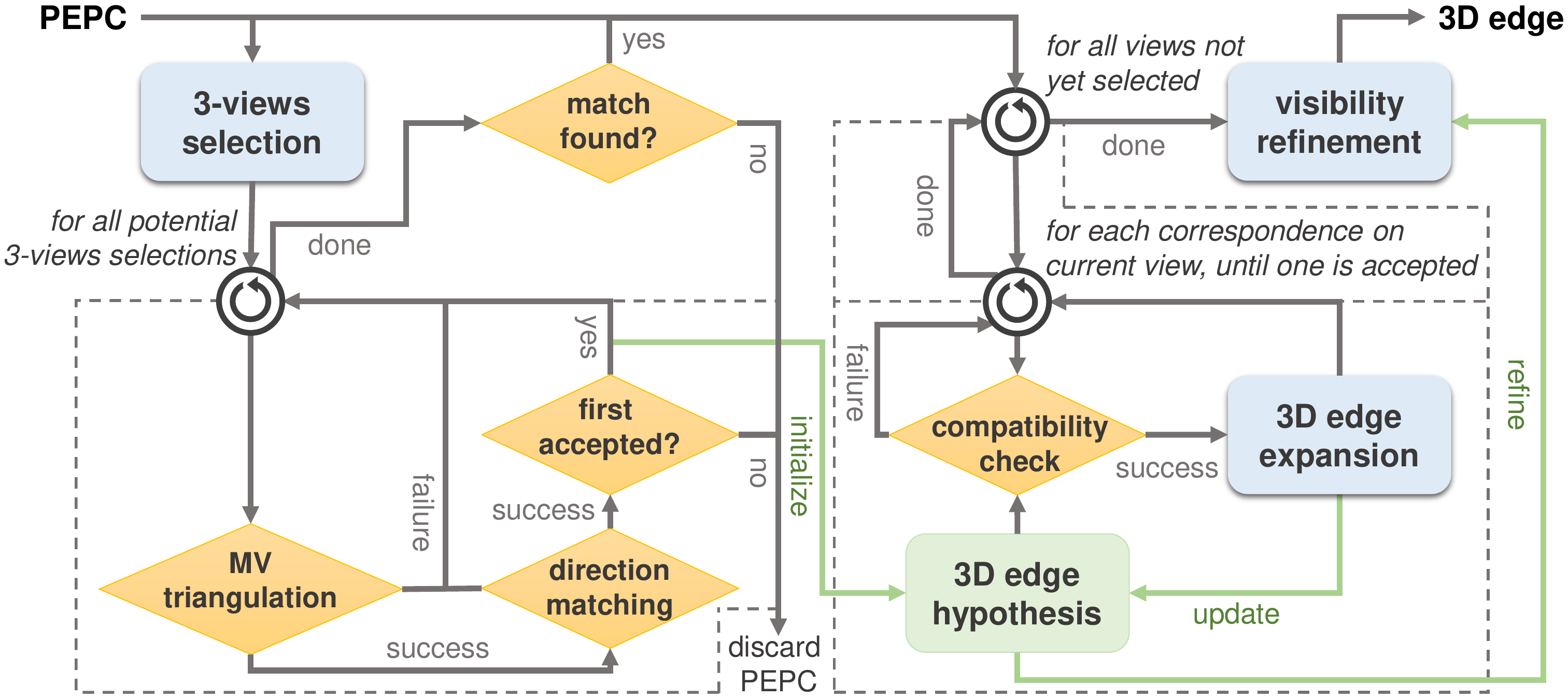}
    \end{center}
    \caption{\pepc validation pipeline}
    \label{fig:pepcval}
  \end{figure*}

\subsubsection{From \pecs}
The second approach (\fref{fig:1b}) makes use of \pecs generated using technique presented in \sref{sec:edge-matching}. 
Formally, a \pec is a set $\defset{\spmss}{\spmsn1, \spmsn2, \dots \spmsn{\amountofviews}}$, which associates to each image $\oimga$ of the $\amountofviews$ views observing the scene, a set $\sopmsa$ of polylines of the corresponding 2D edge-graph involved in the match. 

To generate the \pepcs, let us consider one of the views with a nonempty set of matched polylines set as the initial view $\iimg$. 
We select an edge-point $\ssep$ on one of the matched polylines. Correspondences on each view $\oimga$ can be identified by intersecting the matched polylines on $\oimga$ set with epipolar line $\iepl$ generated by $\ssep$. Repeating the process for all the views produces a set of possibly corresponding 2D edge-points, \ie, a \pepc. The process is repeated for different initial edge-points, obtained by sampling the polylines on an initial view at fixed intervals, to generate multiple \pepcs from a single \pec. 

\subsection{\pepc validation and 3D edge generation}
\label{sec:pepcval}

Each \pepcs generated with the techniques presented so far associates  a set of potential 2D correspondences on images $\oimga \neq \iimg$ to a 2D edge-point $\ssep$ on $\iimg$. Given a \pepc we define:
\begin{Definition}
\textbf{PEPC-Selection}
A \textit{PEPC-selection} is a subset of the \pepc 2D  correspondences such that each image has at most one correspondence.
\end{Definition}
In \sref{sec:combinatorial-selection} we explain how we choose among the vast set of potential selections to recover the 3D edge-point $\eptd$ that generated $\ssep$.
In \sref{sec:exploration} we present the technique we use to recover from  $\eptd$ the corresponding 3D edge.
In \sref{sec:visibility}, we refine the visibility of the generated edges, and we remove outliers. \fref{fig:pepcval} illustrates the complete \pepc validation and 3D edge reconstruction pipeline.

\subsubsection{Correspondences selection}
\label{sec:combinatorial-selection}

A correct PEPC-selection retains, for each view, the one, if it exists, associated to the initial 2D edge-point $\ssep$ on the initial view $\iimg$ used to generate the \pepc. 
The identification of the correct \pepc-selection is therefore a combinatorial problem defined on the search space of all possible selections of edge-point correspondences. A PEPC-selection of 2D edge-point correspondences requires at least three observations to provide minimal geometrical evidence that the correspondences identify the same 3D point, by three-view triangulation. The number of potentially acceptable selections, with at least three views, is therefore:
\feq{
  \xunderbrace[
  \overbrace{\sum\nolimits_{\substack{\sopcsa \in \spcss, \\\sopcsa \neq \sipcs}}\setcardinality{\sospcsa}}^{\substack{\text{pairs of the form} \\ \makepair{\sisep}{\sopca}}} + 
    \xoverbrace{1}^{\substack{\text{only $\sisep$} \\ \text{is selected}}}  
  ]
  {\prod\nolimits_{\substack{\sopcsa \in \spcss, \\\sopcsa \neq \sipcs}} \left( \setcardinality{\sospcsa} + 1 \right)}_{\substack{\text{all combinations} \\ \text{that include $\sisep$}}}
  -
  \xunderbrace{
    \overbrace{\sum\nolimits_{\substack{\sopcsa \in \spcss, \\\sopcsa \neq \sipcs}}\setcardinality{\sospcsa}}^{\substack{\text{pairs of the form} \\ \makepair{\sisep}{\sopca}}} -
    \xoverbrace[\sum\nolimits_{\substack{\sopcsa \in \spcss, \\\sopcsa \neq \sipcs}}\setcardinality{\sospcsa}]{1}^{\substack{\text{only $\sisep$} \\ \text{is selected}}}  
}_{\substack{\text{invalid combinations with} \\ \text{less than 3 selected correspondences}}}\spcomma
}
\noindent
where $\sopcsa$ is the set of correspondences in the \pepc on view $\oimga$, and $\spcss$ is the set of all $\sopcsa$. To reduce the size of the search space, we initially limit the selection problem to only three views, of which one is the initial view $\iimg$. The other two views $\oimga$ and $\oimgb$ can be chosen arbitrarily. All the potential selections on the three views, amounting to $\setcardinality{\sopcsa}*\setcardinality{\sopcsb}$, can be independently checked for correctness. 

A PEPC-selection  is not acceptable if it is not possible to generate a 3D point from it, through multi-view triangulation, with a small maximum reprojection error of $\epsilon$ (\eg, $\epsilon~\approx~2-3$ \emph{px}). Due to inaccuracies, however, incorrect PEPC-selections can  satisfy the geometrical constraint imposed by the triangulation, hence to be considered valid a selection must:
\begin{enumerate*}[label=(\roman*)]
  \item triangulate to a valid 3D point
  \item generate a valid 3D edge, as presented in \sref{sec:exploration}
\end{enumerate*}
. We only accept a PEPC-selection if it is the only one in the \pepc that respects the above conditions. 

\subsubsection{3D Edge reconstruction}
\label{sec:exploration}

Given an initial PEPC-selection and the triangulated 3D edge-point we check whether it is possible to follow the 2D edge-graphs among images to generate a sequence of  corresponding 2D edge-points. 
Starting from the initial edge-point $\iiep$ generated by a 3D edge-point $\iep$ on the first view $\iimg$, it is possible to follow the 2D polyline in two different directions. 
The first step match the different directions of the polylines on only three images, \ie, $I_0$, $I_1$ and $I_2$.
We first move along the corresponding polyline in the direction $d_0$, by a small fixed length $\matchdirectionfollowlength~\approx~10$ \emph{px}, and select a new edge-point $\oaiep$ on $\iimg$. The sampling interval $\matchdirectionfollowlength$ has been chosen, without tuning, to closely follow the direction changes of a polyline, without generating irrelevant edge-points.
The correspondences of $\oaiep$ on the other view $\oimga$ can be identified by tracing the epipolar line on $\oimga$  and selecting the first intersection $\oaoepa$ between that line and the corresponding polyline on $\oimga$, starting from the initial edge-point observation $\ioepa$, and moving towards the direction $d_i$. 
We apply this process on both views $1$ and $2$, and we verify that the two edge-point correspondences, $\oaoepa$ and $\oaoepb$ together with $\oaiep$ triangulate to a valid 3D point.
If this operation is successful, the direction is considered valid. New edge-points can be recursively found following the directions $(d_0, d_1, d_2)$ and the opposite $(-d_0, -d_1, -d_2)$ until a failure, either in finding correspondences or in the multi-view triangulation, occurs. 
Note  that the extent of the movement between edge-point samples on the first view controls for the degree of approximation of curved edges.

The above process is the initial step in the definition of a new 3D edge. If a solution is accepted for three views, \ie, it is the only valid selection, other potential 2D edge-point correspondences in the \pepc on additional views, excluded in the initial selection, should still be integrated if compatible with the current 3D edge. 
We consider a new potential 2D to 3D edge-point observation $\sopccn1$ for the initial 3D edge-point $\iep$, on a new view $\oimgc$. The first step in validating the new observation is checking whether it is compatible, by multi-view triangulation, with the current bidimensional observations of $\iep$ and that it is possible to match directions between the current 3D polyline-edge, and the 2D polyline on $\oimgc$. Using the new observations, we can further extend the current 3D edge, and improve its accuracy.

\begin{table*}[t]
\normalsize
\centering
  \caption{Comparison between \sysname and OpenMVG.}
  \label{tab:expRes}
\begin{tabular}{lrlcccccc}
      &   & num.& \multicolumn{3}{c}{point cloud} & \multicolumn{3}{c}{mesh} \\ \cmidrule(lr){4-6} \cmidrule(lr){7-9}
      &   & points & MAE   & RMSE & $\sigma$ & MAE   & RMSE & $\sigma$ \\
\hline
\multirow{2}{*}{fountain-P11}&
OpenMVG   & 5570 & \textbf{8.433} & \textbf{9.603} & \textbf{12.78} & 88.94 & 209.3 & 189.4 \\
& \sysname  & \textbf{41725} & 12.19 & 15.98 & 20.10 & \textbf{64.58} & \textbf{159.3} & \textbf{145.6}  \\
\hline
\multirow{2}{*}{DTU-006}&
OpenMVG   & 5903 & \textbf{0.477} & \textbf{0.938} & \textbf{1.052} & 4.077 & 11.53 & 10.79\\
& \sysname  & \textbf{47927} & 0.542 & 1.230 & 1.344 & \textbf{2.805} & \textbf{8.497} & \textbf{8.021}\\
\hline
\multirow{2}{*}{DTU-023}&
OpenMVG   & 9651 & 0.826 & \textbf{1.886} & 2.059  & 4.585 & 9.223 & 8.003\\
& \sysname & \textbf{97770}  & \textbf{0.825} & 1.911 & \textbf{0.020}  & \textbf{3.207} & \textbf{7.898} & \textbf{7.218}\\
\hline
\multirow{2}{*}{DTU-028}&
OpenMVG   & 5008 & 1.961 & 3.607 & 4.106  & 20.24 & 54.07 & 50.14\\
& \sysname & \textbf{46220} & \textbf{1.013} & \textbf{2.766} & \textbf{2.946}  & \textbf{13.22} & \textbf{35.74} & \textbf{33.21}\\
\hline
\multirow{2}{*}{DTU-037}&
OpenMVG   & 4321 & \textbf{1.326} & \textbf{1.830} & \textbf{2.260}  & \textbf{23.36} & 40.21 & 32.72\\
& \sysname & \textbf{38577} & 1.478 & 2.372 & 2.795  & 25.32 & \textbf{40.12} & \textbf{31.12}\\
\hline
\multirow{2}{*}{DTU-098}&
OpenMVG  & 2091 & 4.658 & 9.501 & 10.58  & 24.89 & 55.32 & 49.40\\
& \sysname & \textbf{26575} & \textbf{3.831} & \textbf{8.449} & \textbf{9.277}  & \textbf{5.907} & \textbf{18.42} & \textbf{17.45}\\
\hline
\multirow{2}{*}{DTU-118}&
OpenMVG  & 1839 & \textbf{2.770} & \textbf{7.311} & \textbf{7.818}  & 19.93 & 39.49 & 34.10\\
& \sysname & \textbf{14611} & 3.065 & 7.894 & 8.468  & \textbf{7.806} & \textbf{21.06} & \textbf{19.55}\\
\hline
\bfseries average variation & & & +2.8\% & +13.7\% & +10.6\% & \textbf{-36.0\%} & \textbf{-30.3\%} & \textbf{-29.2}\%\\
\end{tabular}
\end{table*}

\subsubsection{Visibility refinement and outliers filtering}
\label{sec:visibility}

Once a 3D polyline-edge $\spltd$ is generated, we can optimize its visibility information by checking whether the 3D edge is visible in a view that was not considered so far. Let us consider $\oimga$, on which no observation of the 3D polyline $\spltd$ has been found. 
New bidimensional observations of polyline $\spltd$, if existent on $\oimga$, are expected to be near the projection of $\spltd$ on $\oimga$. In the proposed system, we look for new 2D edge-point observations for each of the 3D edge-points that define $\spltd$. Let us consider $\oepa \in \spltd$, and its projection $\projectpt{\oepa}{\indexa}$ on $\oimga$. The goal is finding, if existent, a new edge-point observation $\oaoepa$ of $\oepa$ on $\oimga$. We look for polylines within a distance $\expandvisibilitymaxpldist$ from $\projectpt{\oepa}{\indexa}$. If a single polyline $\spla$ is found, we select the edge-point on $\spla$ closest to $\projectpt{\oepa}{\indexa}$ as the new potential 2D observation $\oaoepa$ of $\oepa$. By multi-view triangulation we verify the compatibility of $\oaoepa$ with the current observations set of $\oepa$. If this check is successful, we can verify that the 2D polyline $\spla$ is compatible, in the vicinity of $\oaoepa$, with the 3D polyline $\spltd$. This can be done by matching both directions of $\spltd$ from the initial point $\oepa \in \spltd$, with the two directions on $\spla$ starting from $\oaoepa$, using techniques analogous to the ones presented in \sref{sec:exploration}. This process is applied to all edge-points of $\spltd$ that have not  been observed on $\oimga$ yet, and can be repeated for every view $\oimga$ to ensure that all potential observations of the 3D polyline $\spltd$ are correctly identified. 

Finally, we consider polylines with a low amount of observations to be likely outliers, hence we filter them out. The minimum amount of observations $\mininlierobservations$ is computed as $\mininlierobservations = max\left(4,\frac{\mediannumberobservations}{2}+1\right)$,
where $\mediannumberobservations$ is the median number of observations for all the 3D edge-points recovered by the system. The output of this final step is a set of accurate 3D polyline-edges that can properly represent even curved edges. 

\section{Experimental Results}
\label{sec:exp}

We evaluate the results obtained by \sysname on the fountain-P11 sequence of the EPFL dataset \cite{strecha2008} and on 6 sequences of the DTU dataset \cite{jensen2014large} by reconstructing the 3D edges, extracting a point cloud by finely sampling them, used to reconstruct a mesh using the algorithm in \cite{romanoni15b} which is then compared with the ground truth.
All the tests have been conducted on a Intel i5-3570K quad-core processor (3.80 GHz frequency, 6 MB smart cache) and 8 GB of DDR3 RAM. The values of the parameters of the algorithms presented in \sref{sec:algo}, such as the maximum reprojection error of $\epsilon$, have been chosen to properly represent the geometrical properties associated with them and have not been subject to tuning in our experiments.

\begin{figure*}[t]
\centering
\setlength{\tabcolsep}{0px}
\begin{tabular}{ccccccc}
fountain-P11&
DTU-006&
DTU-023&
DTU-028&
DTU-037&
DTU-098&
DTU-118\\
\hline
\includegraphics[width=0.14\textwidth]{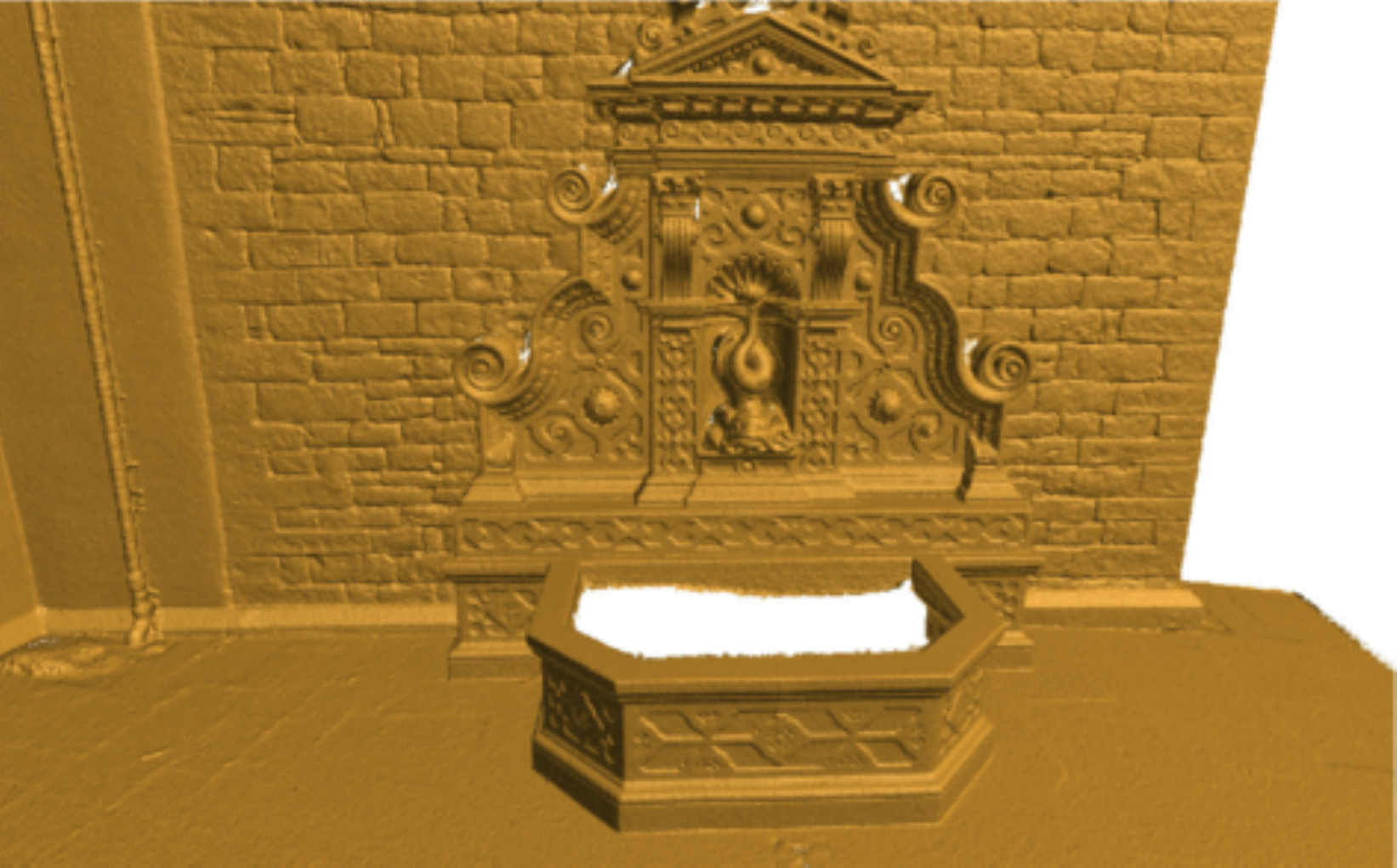}&
\includegraphics[width=0.14\textwidth]{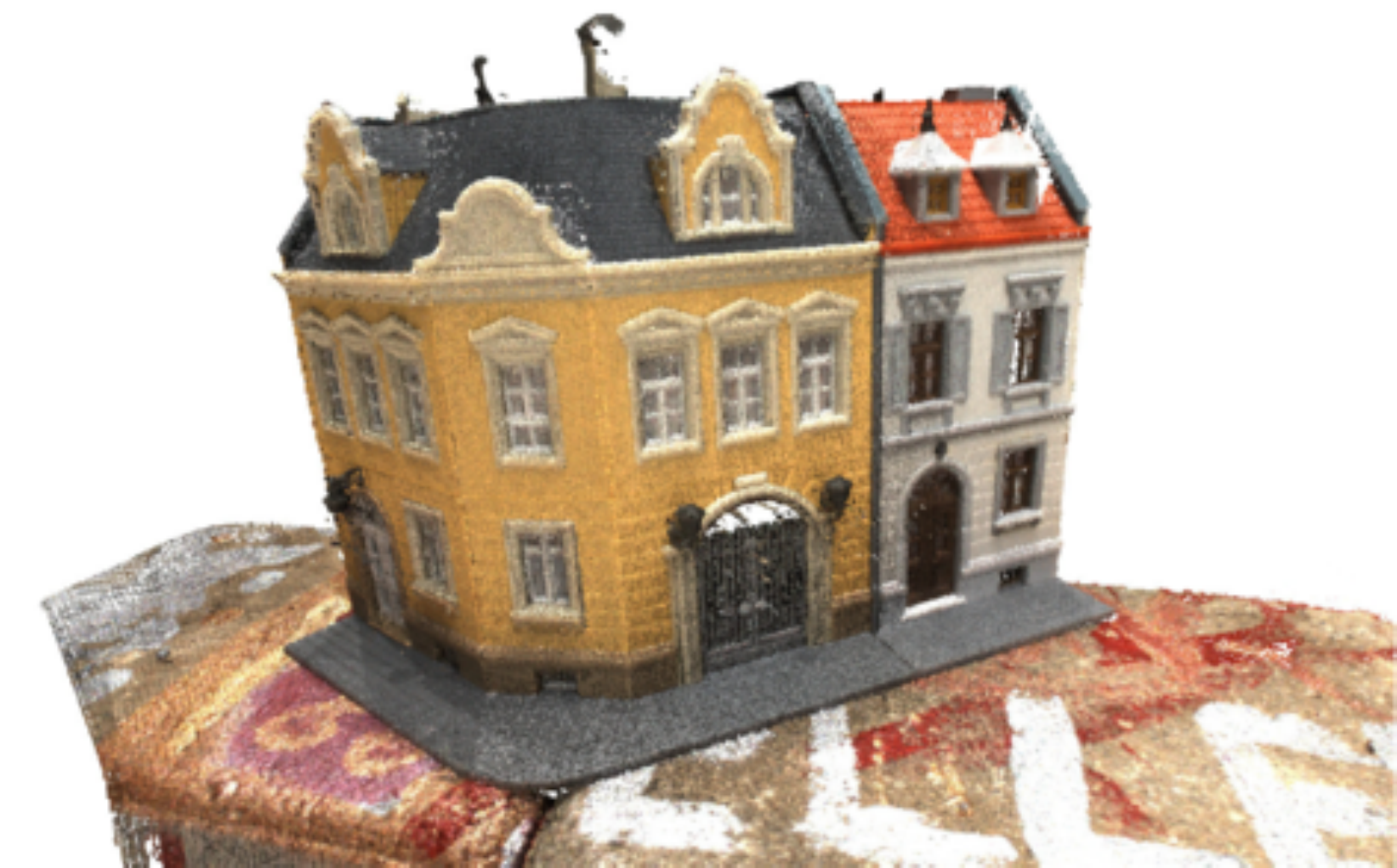}&
\includegraphics[width=0.14\textwidth]{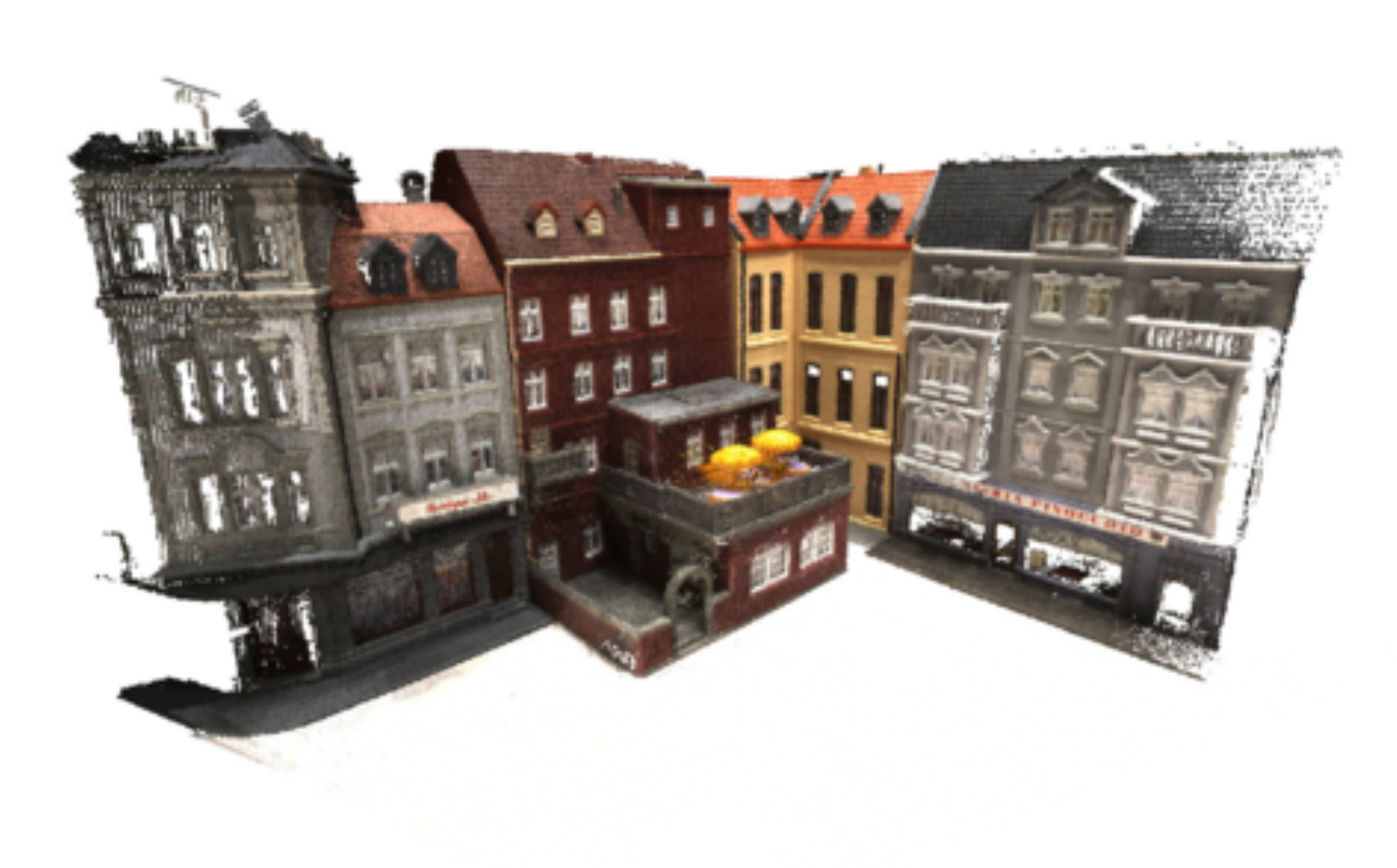}&
\includegraphics[width=0.14\textwidth]{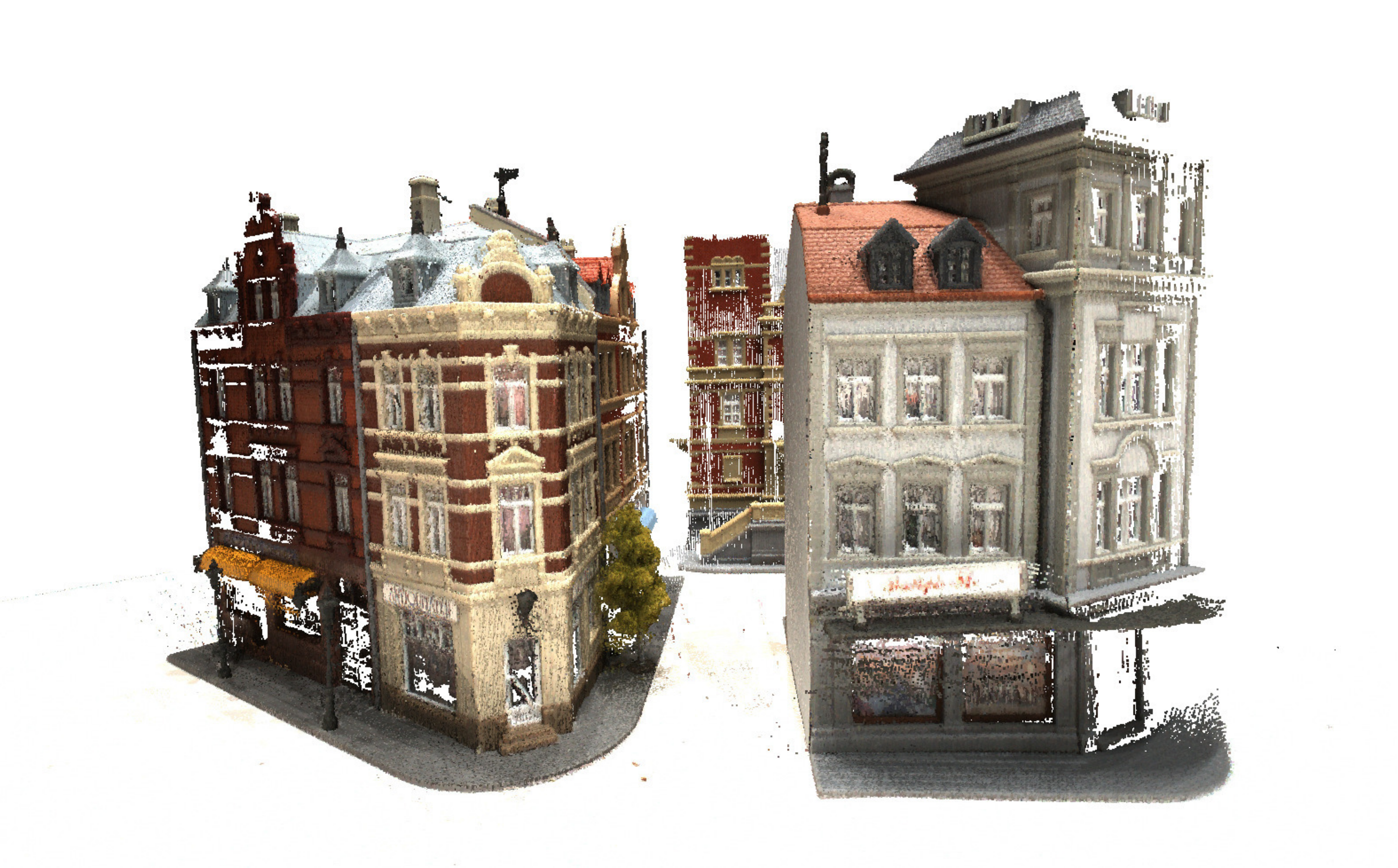}&
\includegraphics[width=0.14\textwidth]{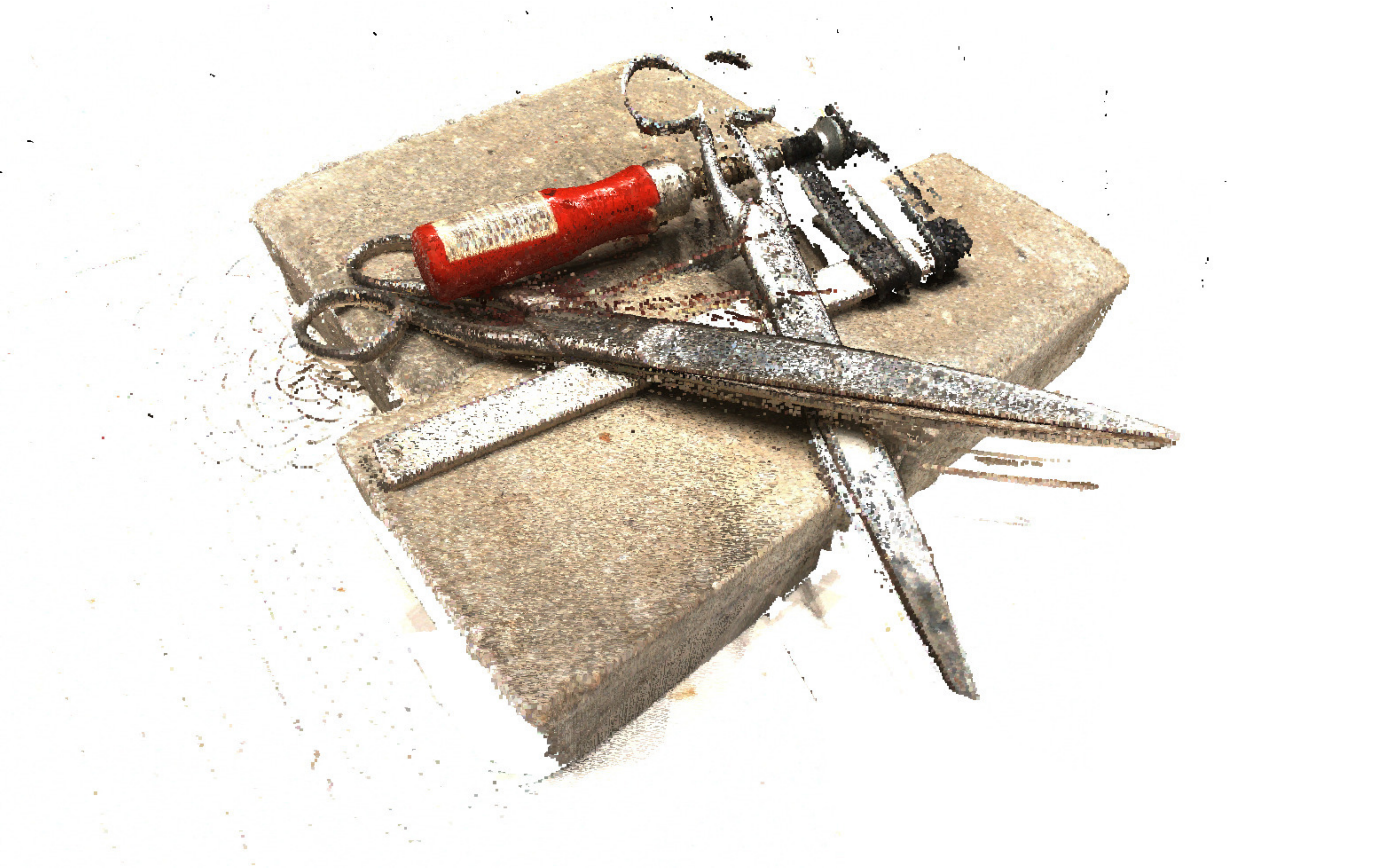}&
\includegraphics[width=0.14\textwidth]{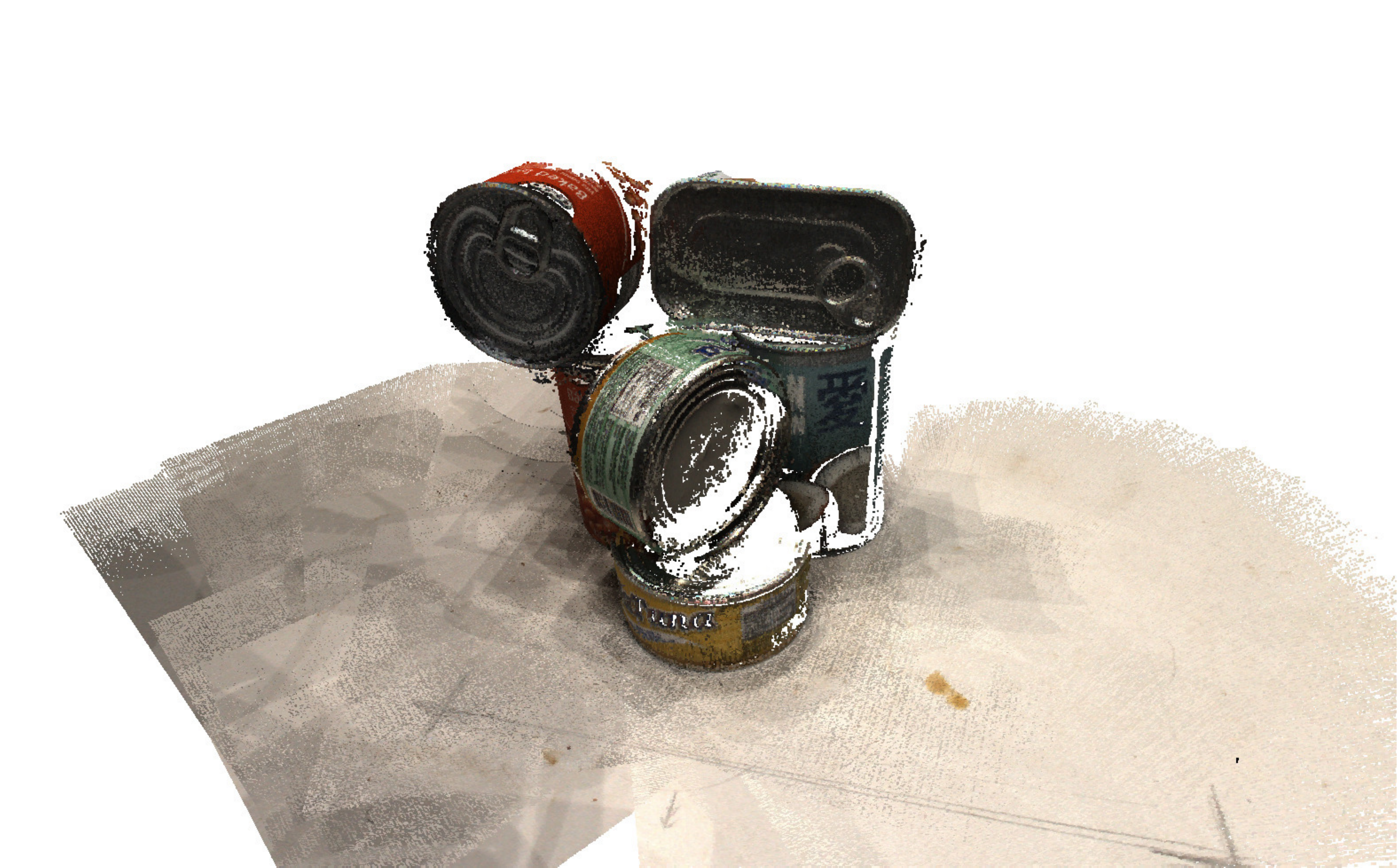}&
\includegraphics[width=0.14\textwidth]{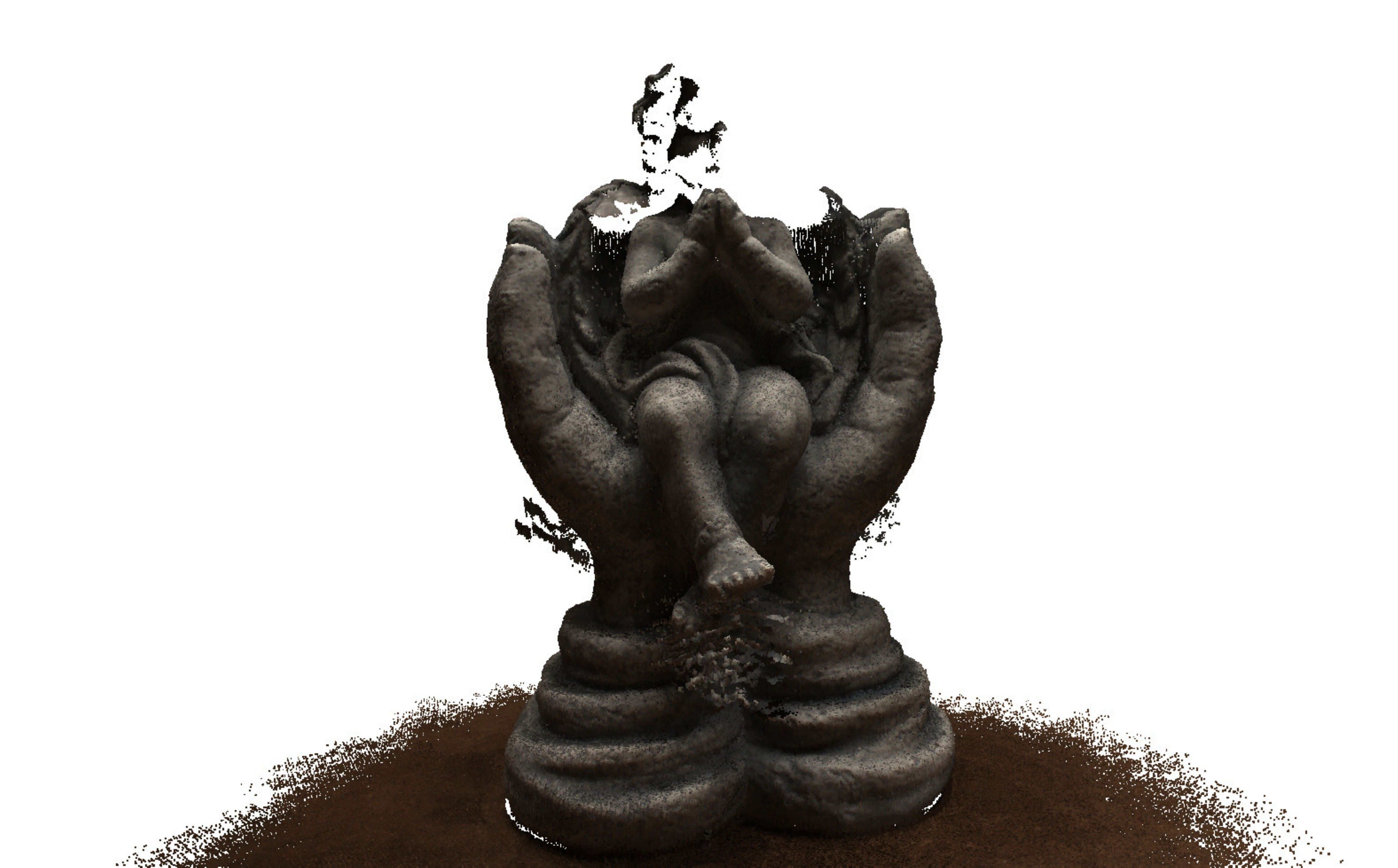}\\
\includegraphics[width=0.14\textwidth]{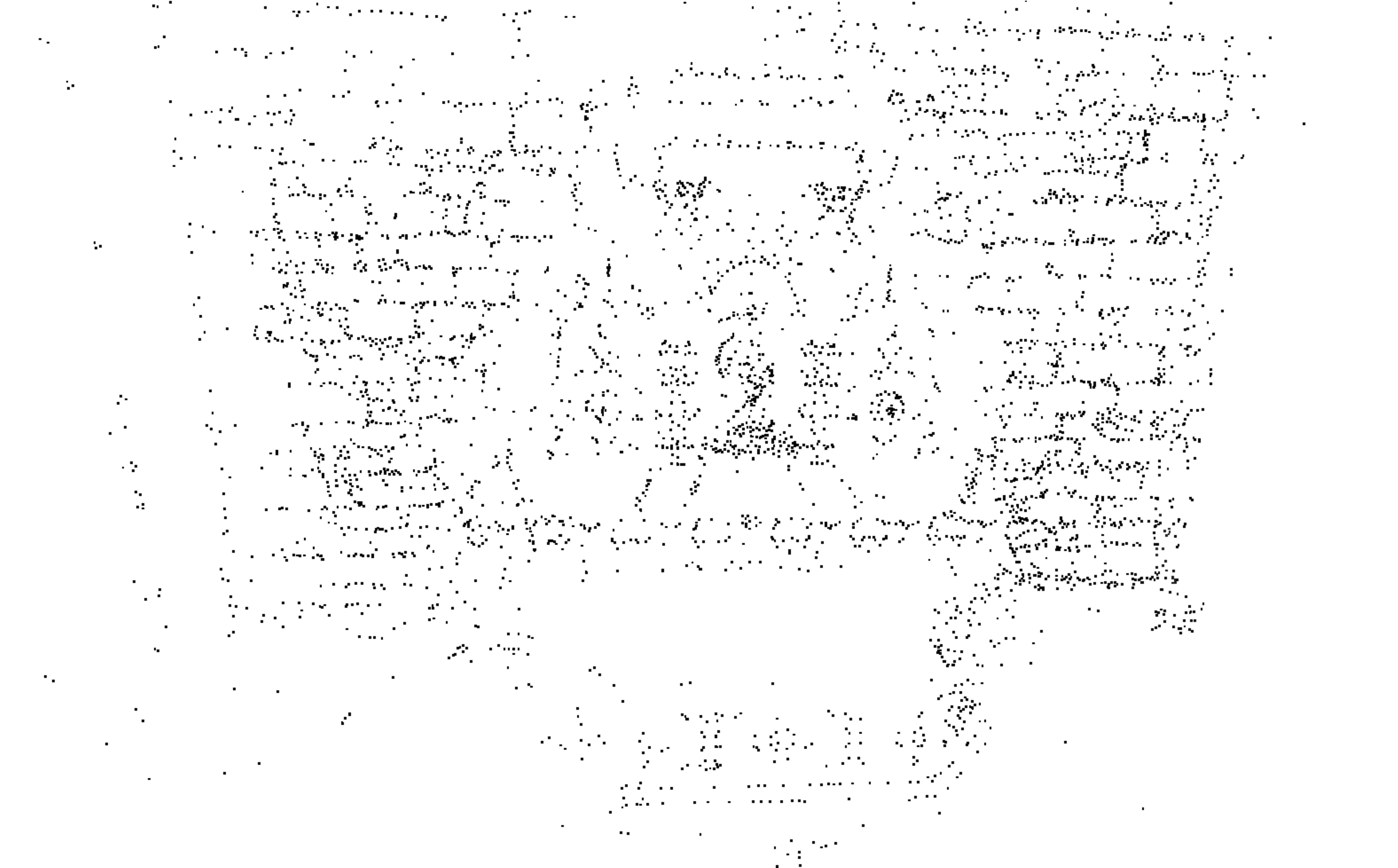}&
\includegraphics[width=0.14\textwidth]{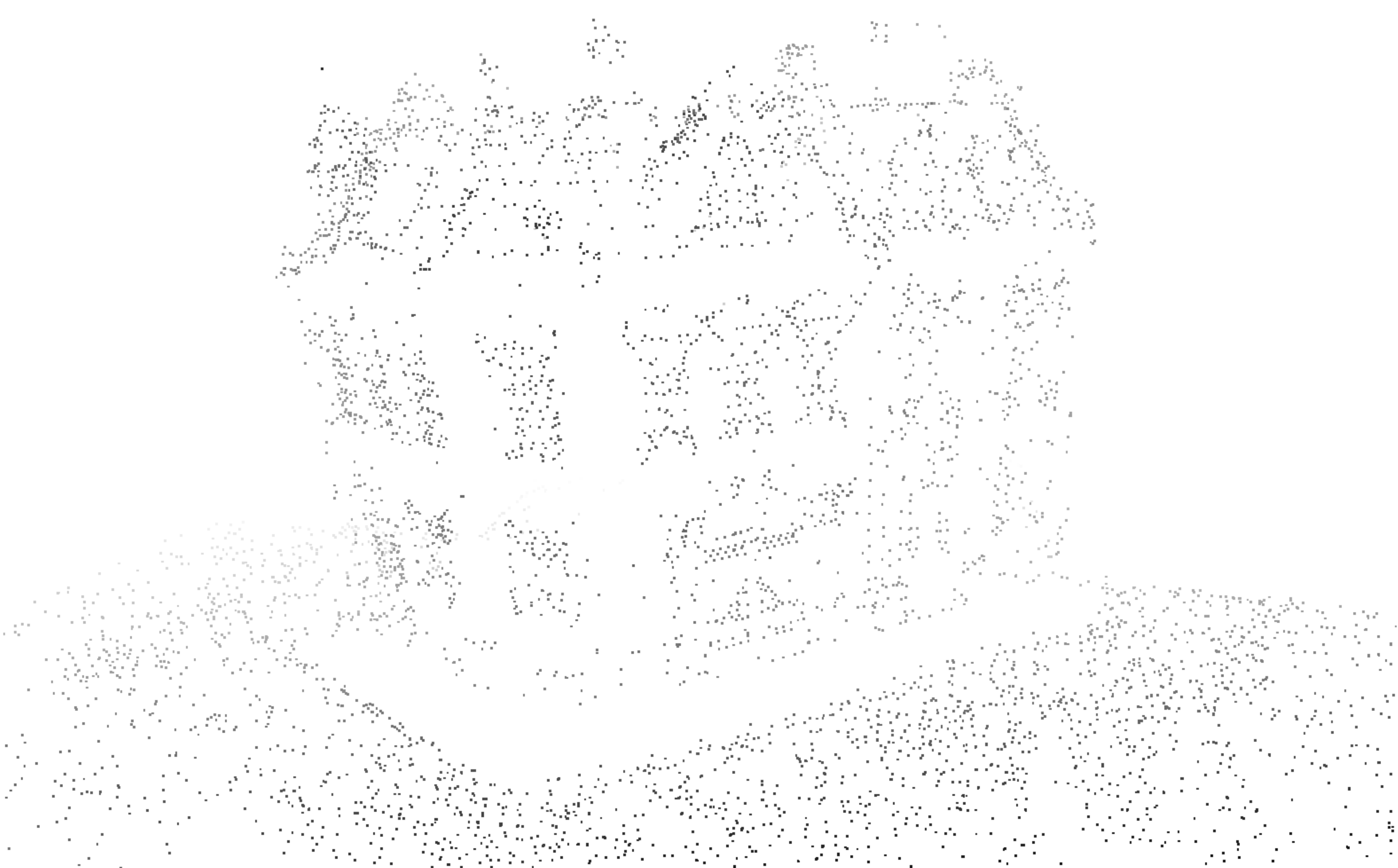}&
\includegraphics[width=0.14\textwidth]{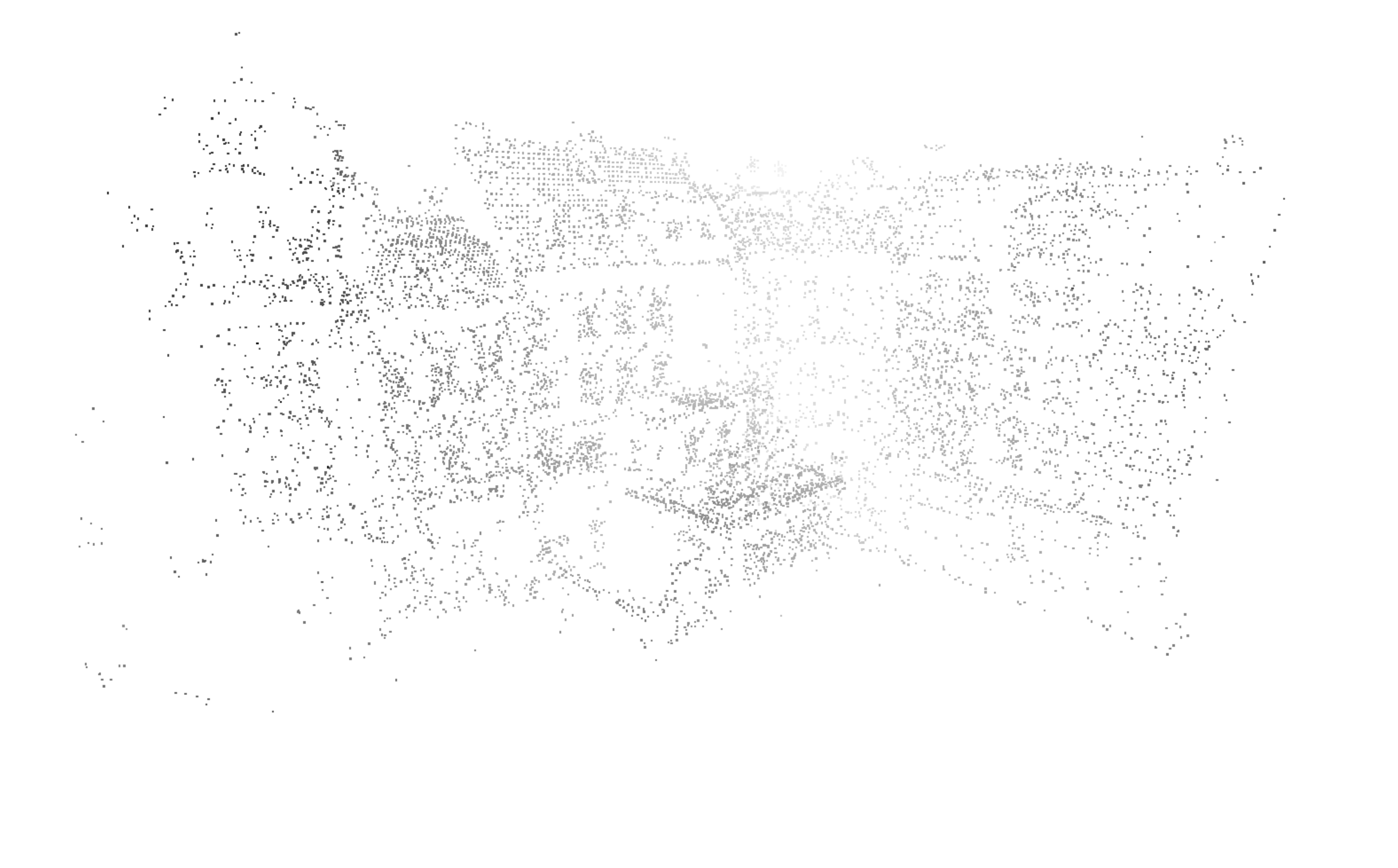}&
\includegraphics[width=0.14\textwidth]{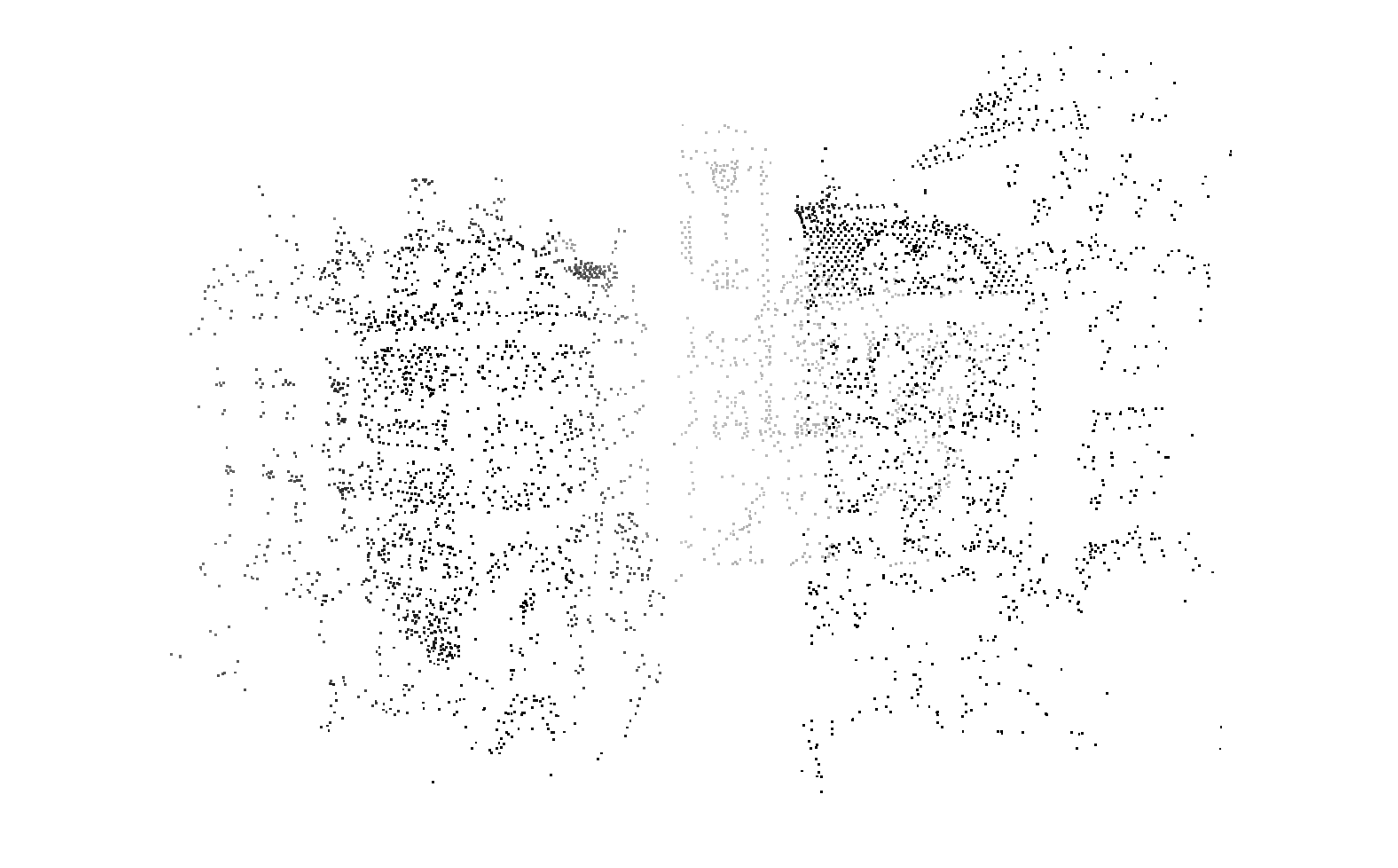}&
\includegraphics[width=0.14\textwidth]{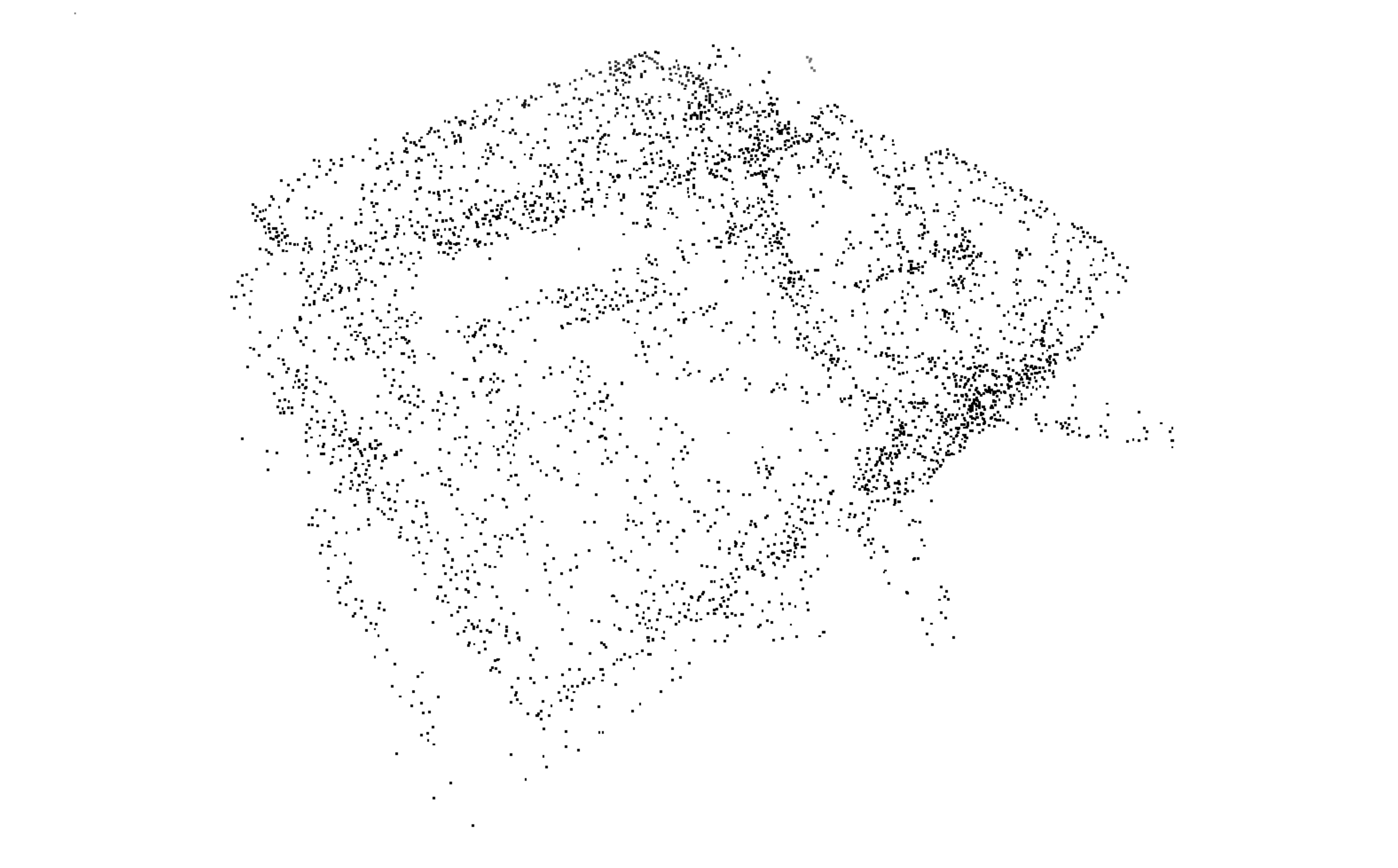}&
\includegraphics[width=0.14\textwidth]{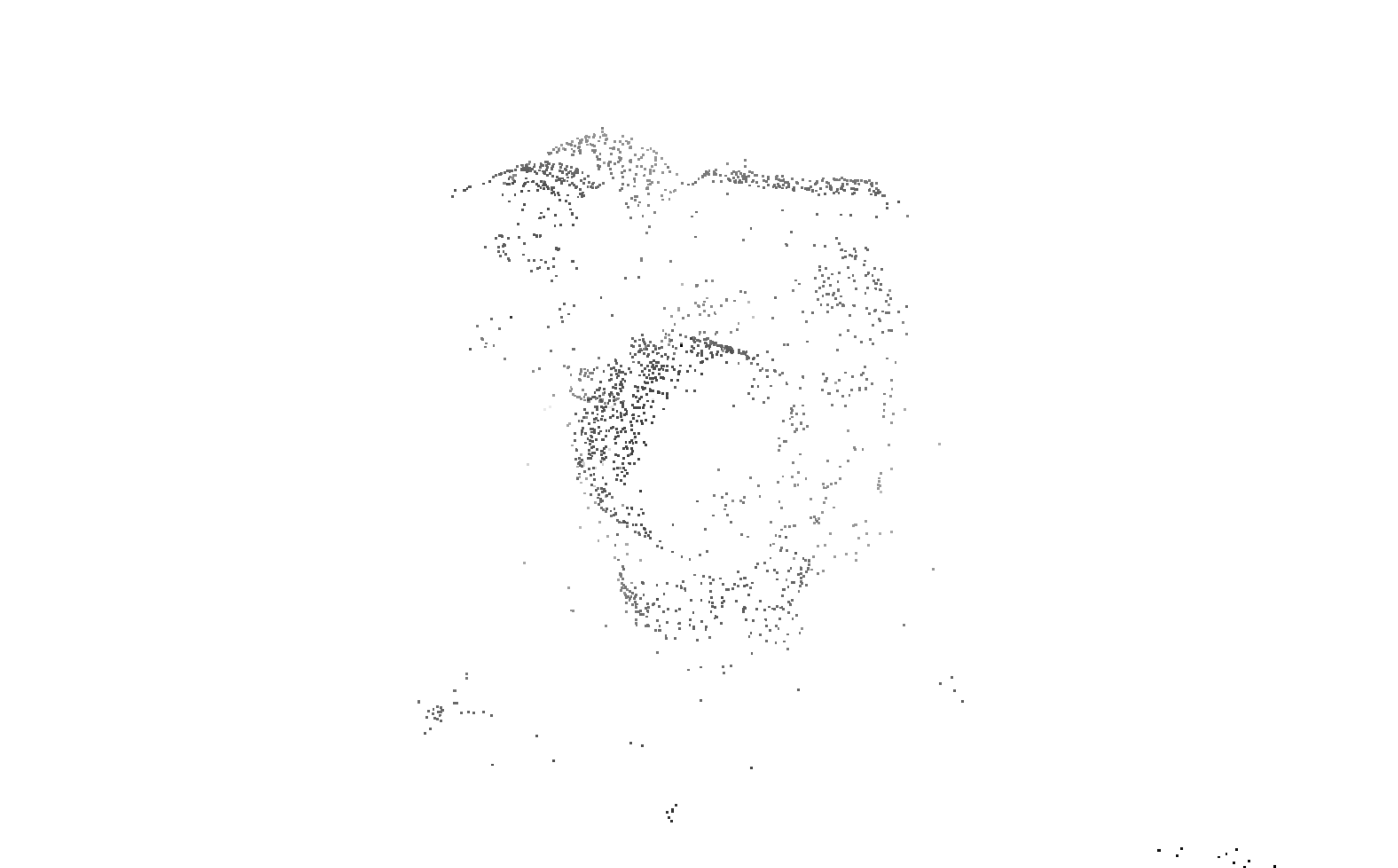}&
\includegraphics[width=0.14\textwidth]{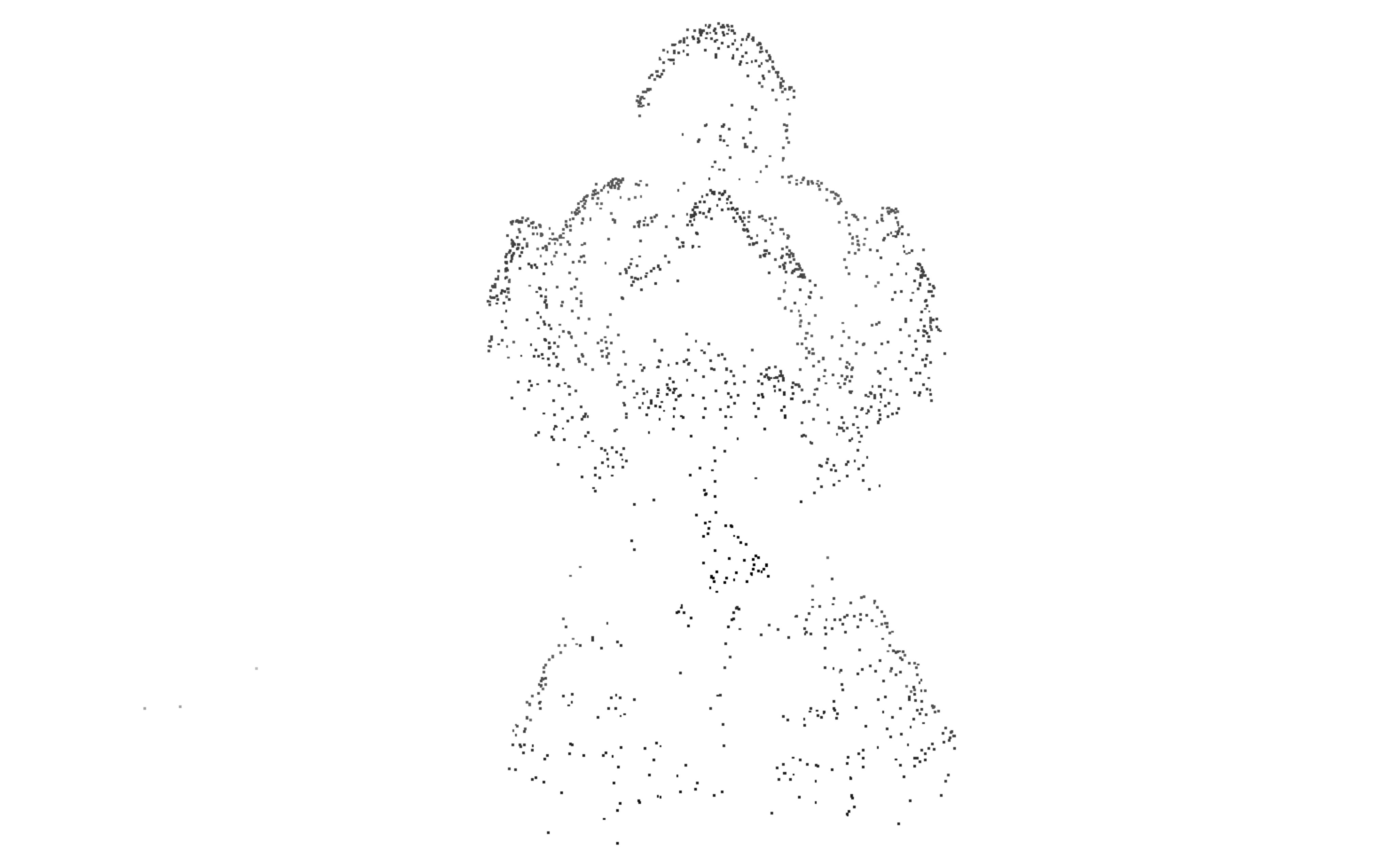}\\
\includegraphics[width=0.14\textwidth]{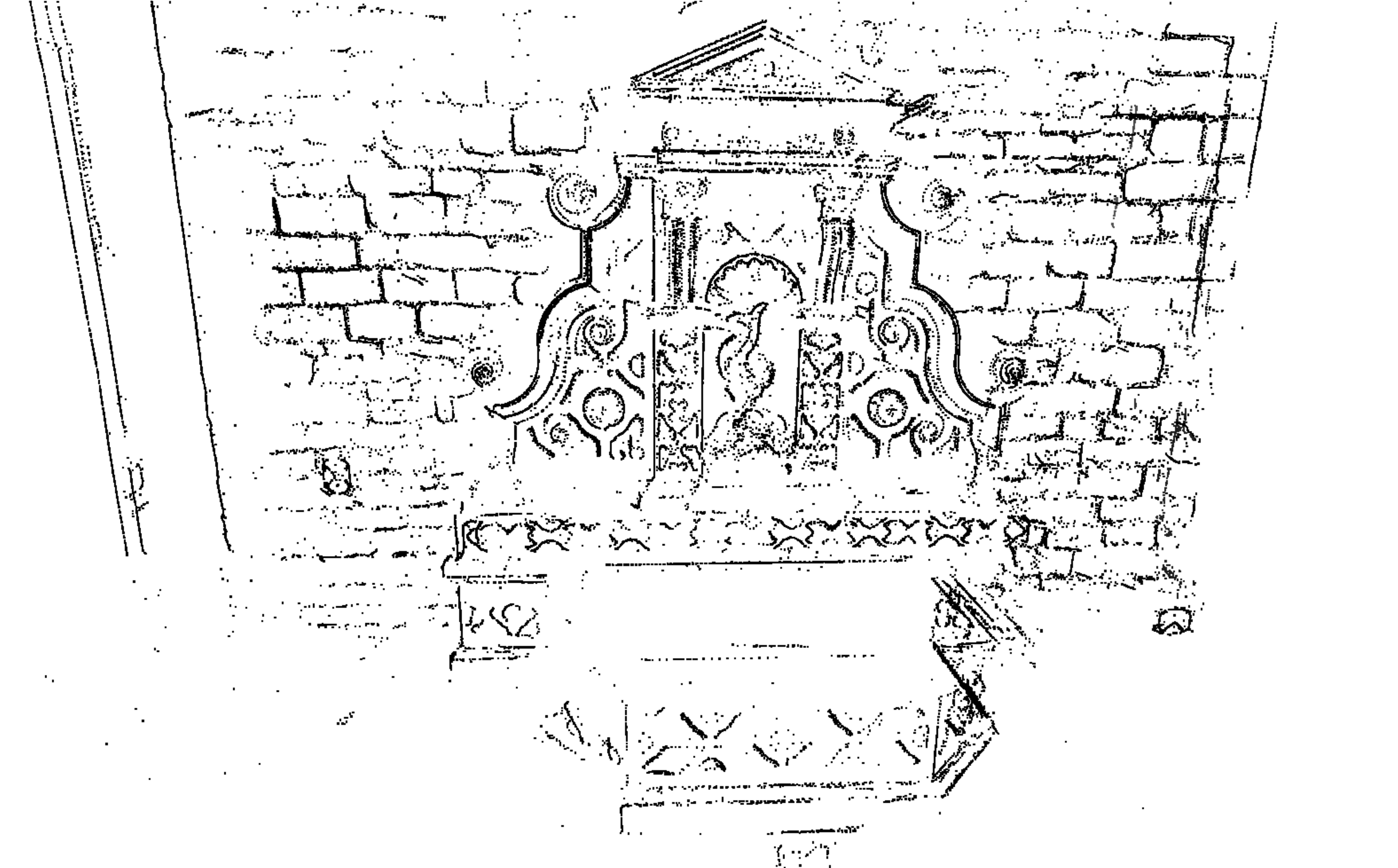}&
\includegraphics[width=0.14\textwidth]{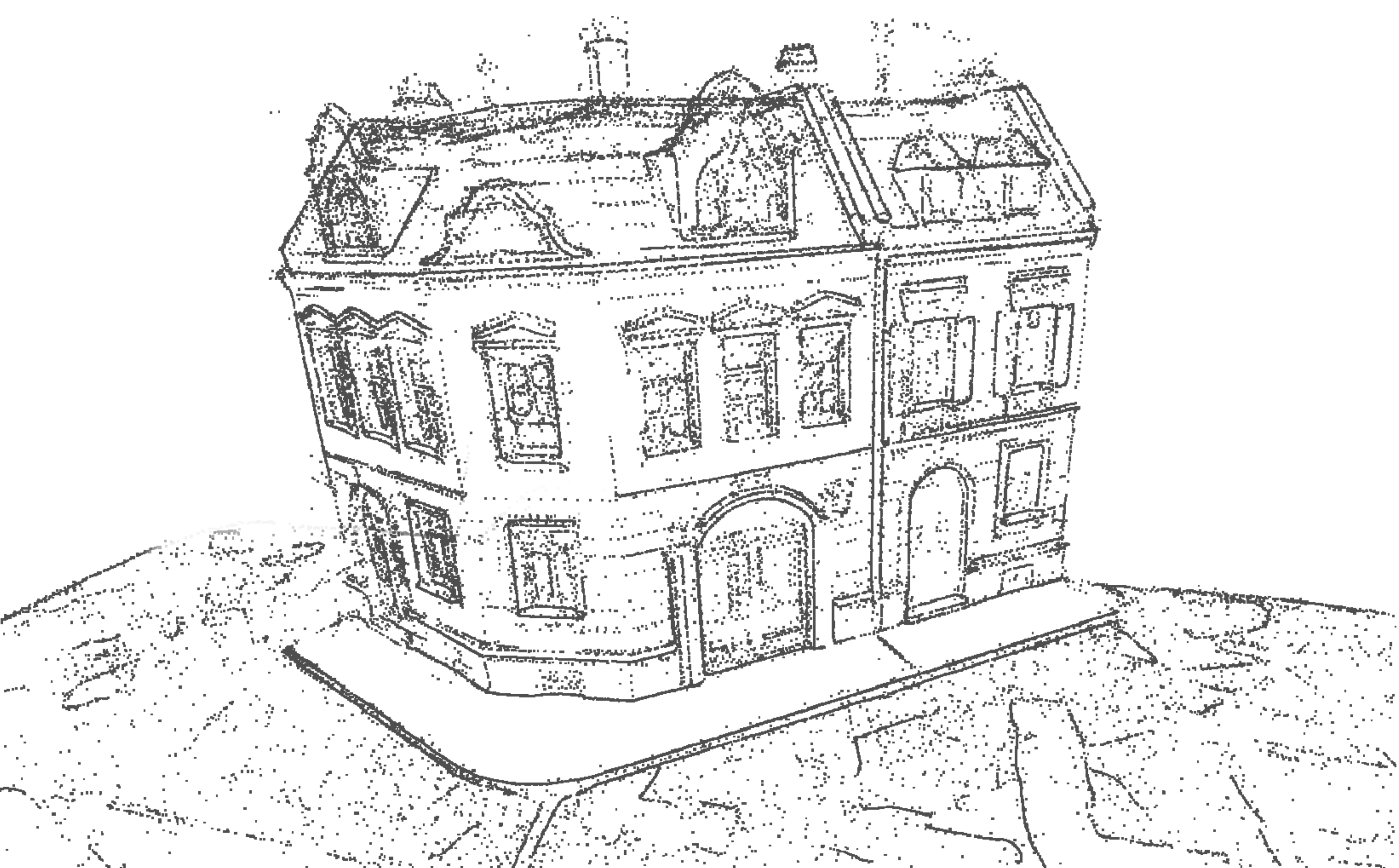}&
\includegraphics[width=0.14\textwidth]{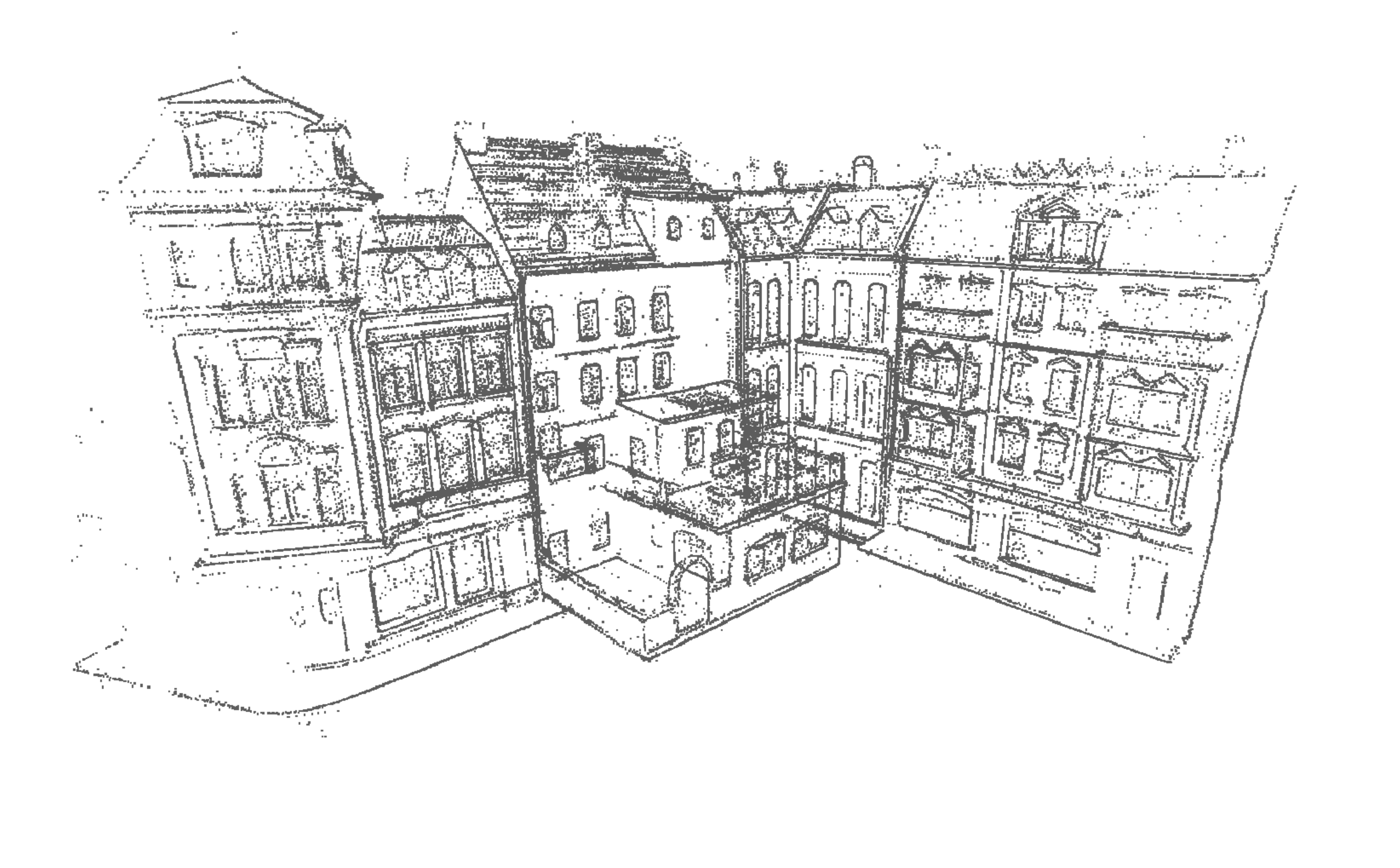}&
\includegraphics[width=0.14\textwidth]{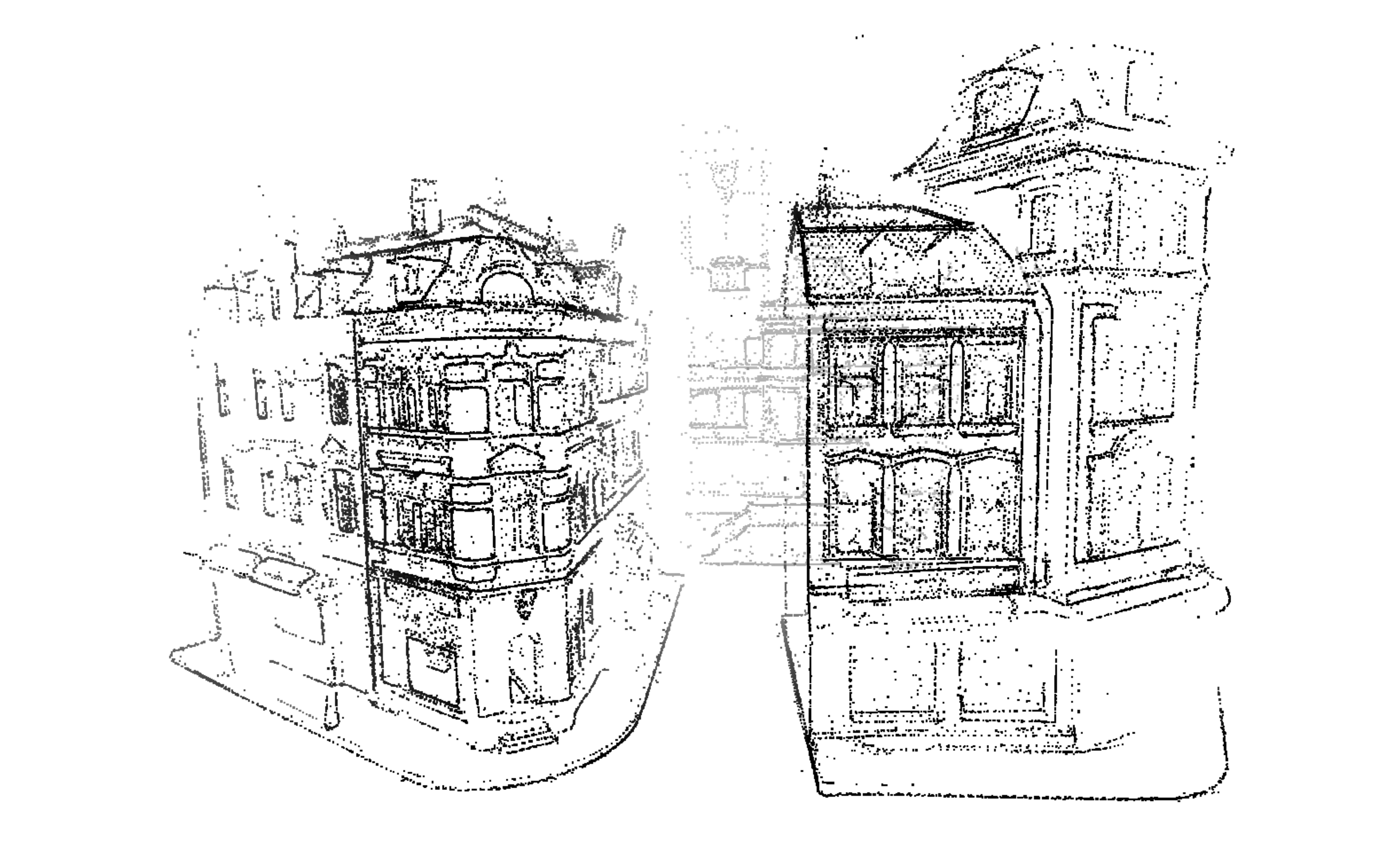}&
\includegraphics[width=0.14\textwidth]{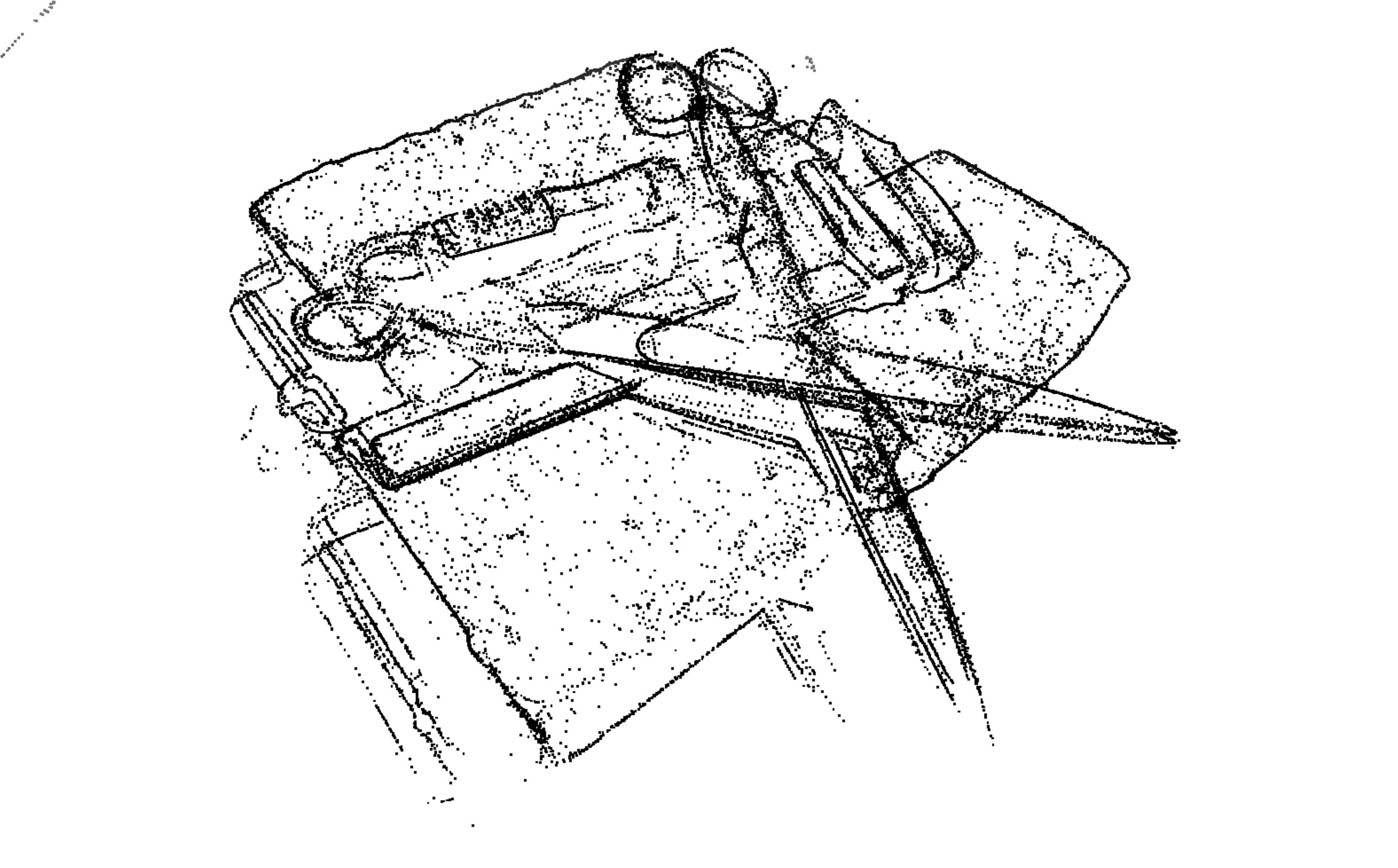}&
\includegraphics[width=0.14\textwidth]{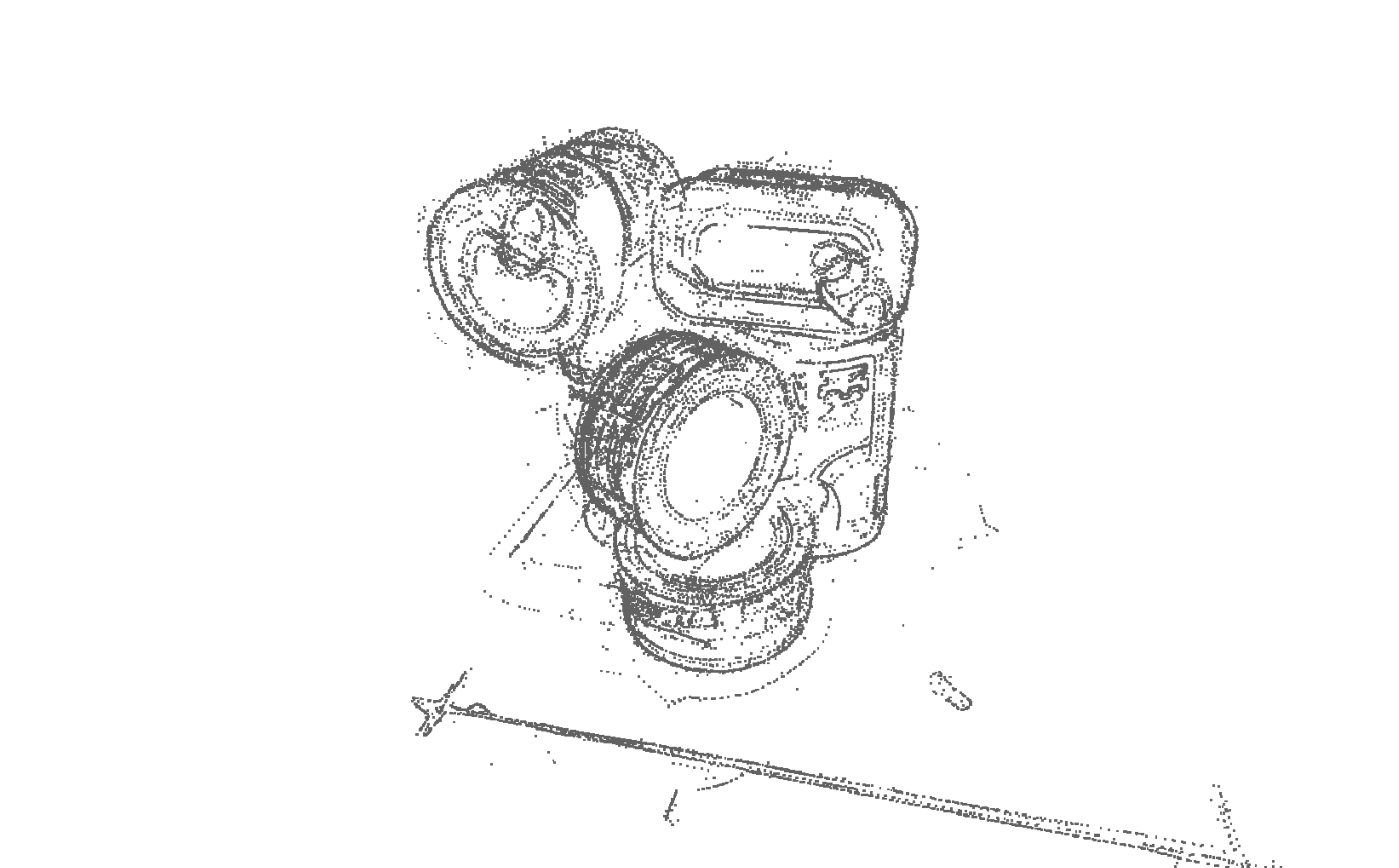}&
\includegraphics[width=0.14\textwidth]{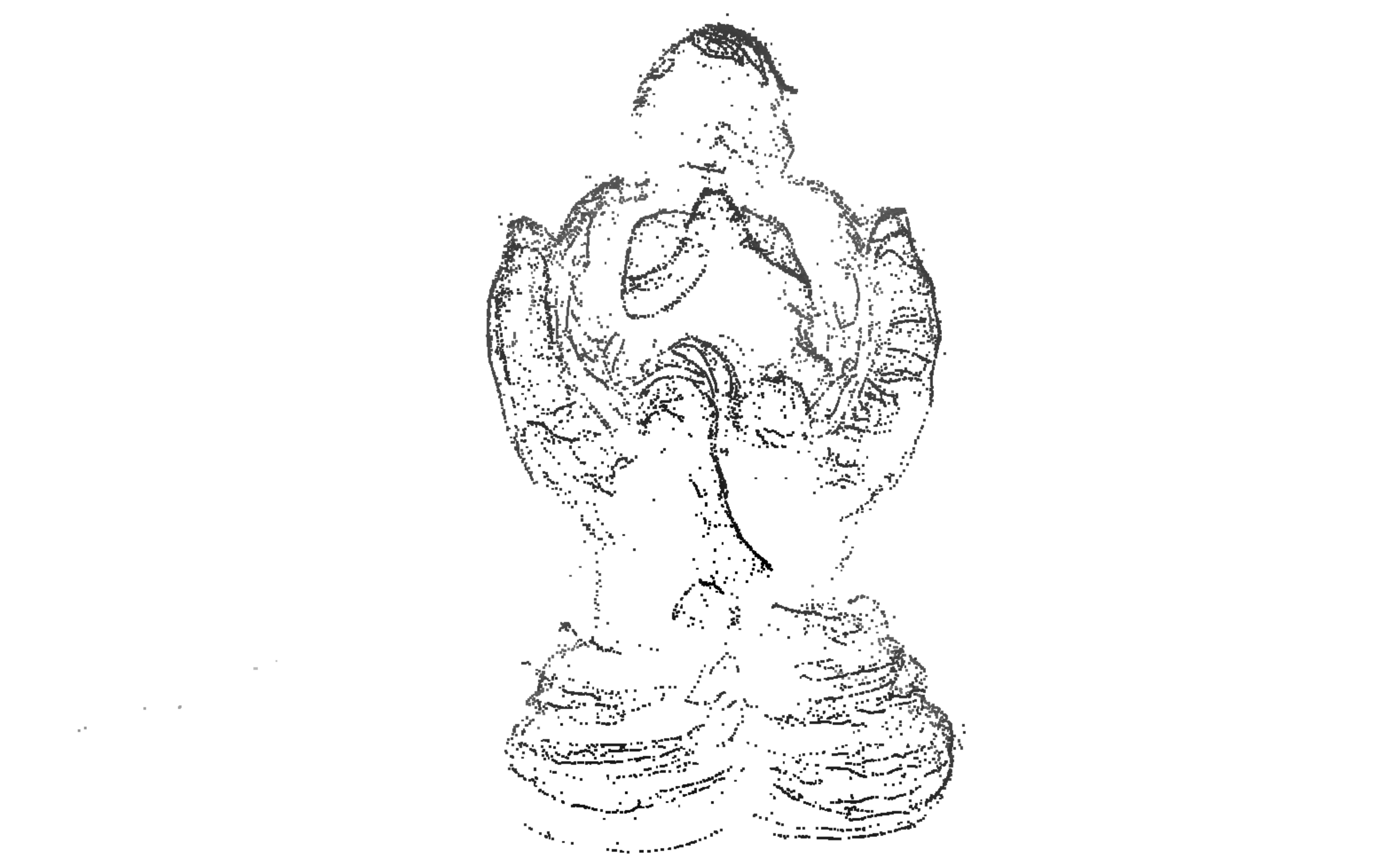}\\
\includegraphics[width=0.14\textwidth]{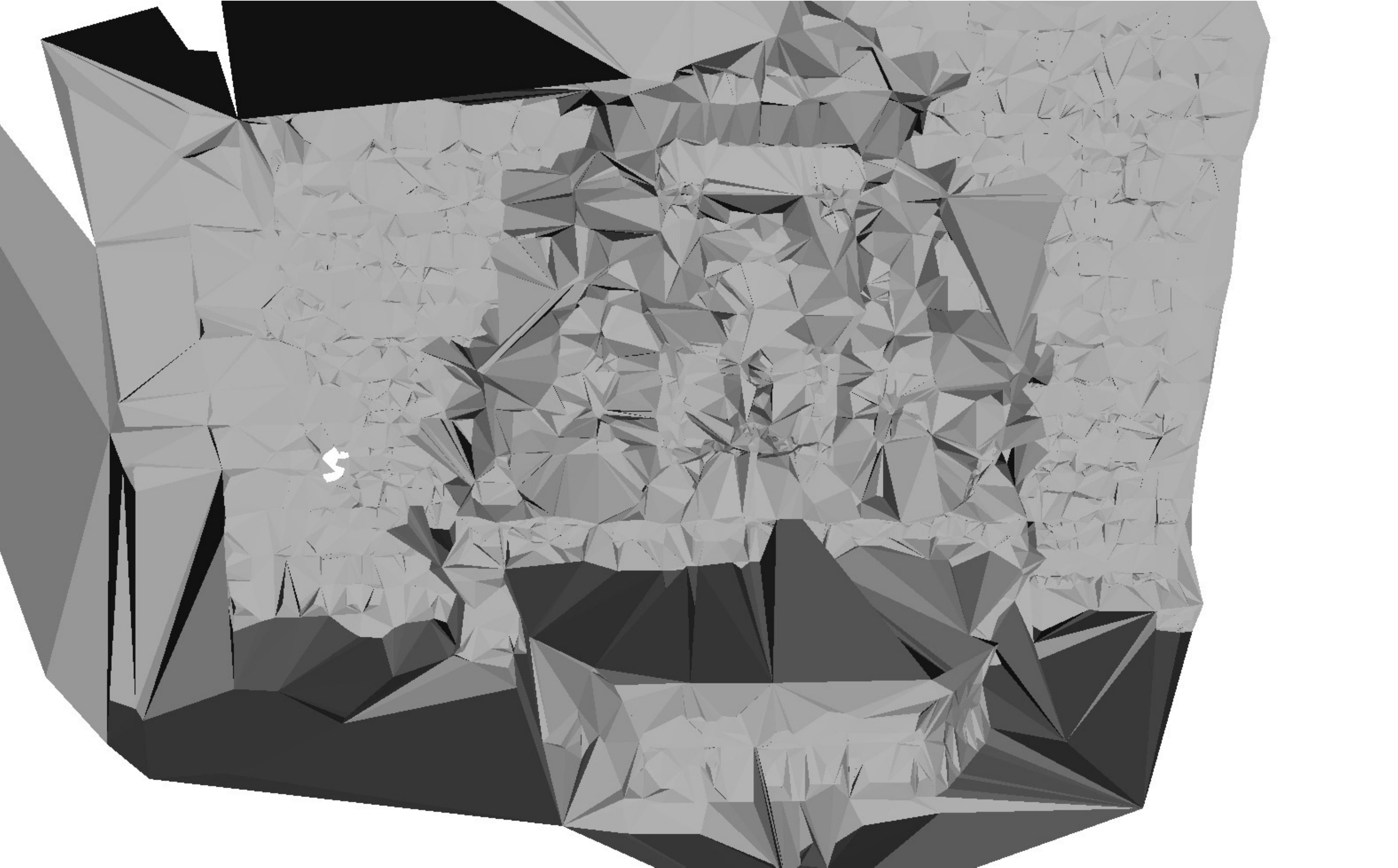}&
\includegraphics[width=0.14\textwidth]{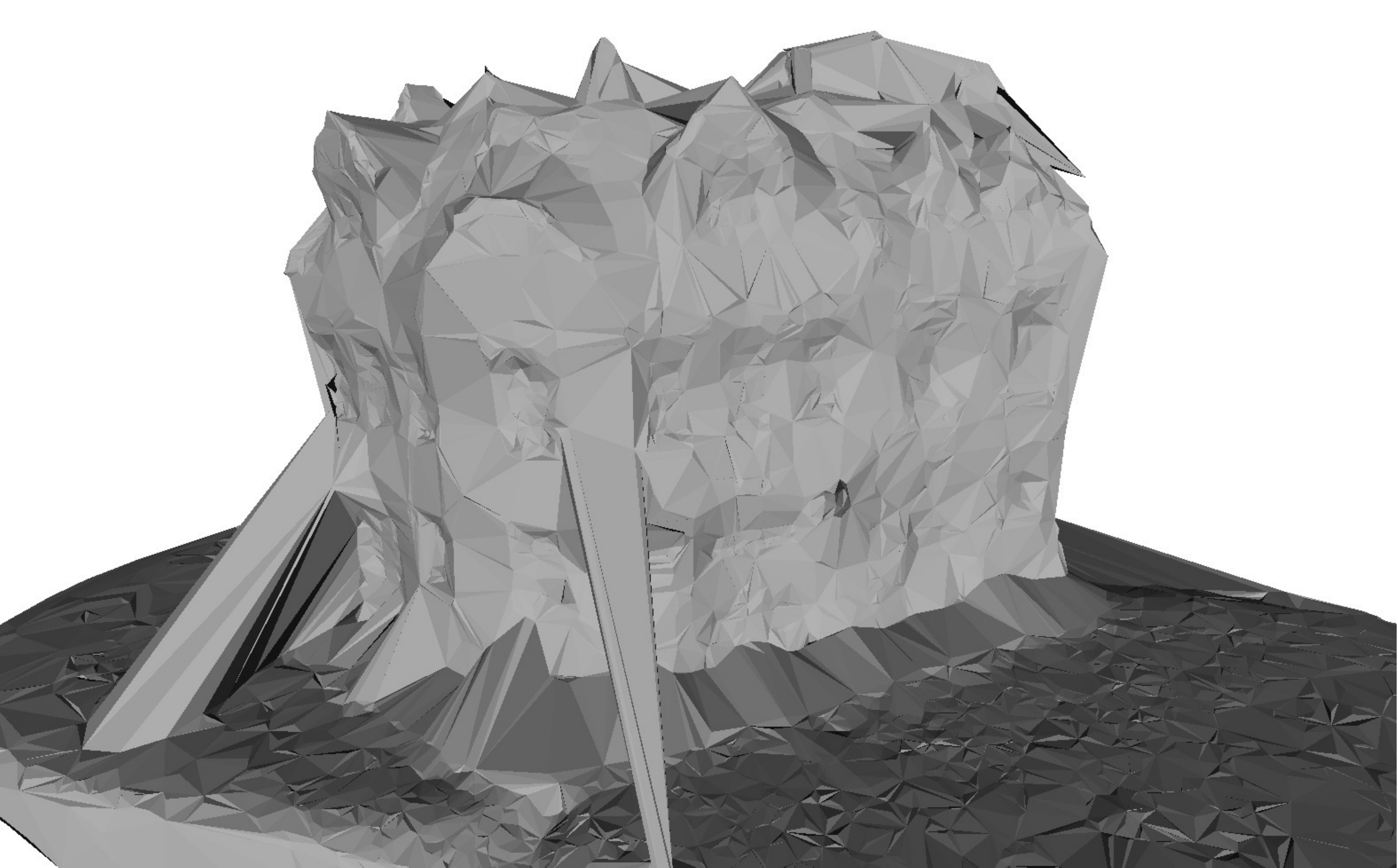}&
\includegraphics[width=0.14\textwidth]{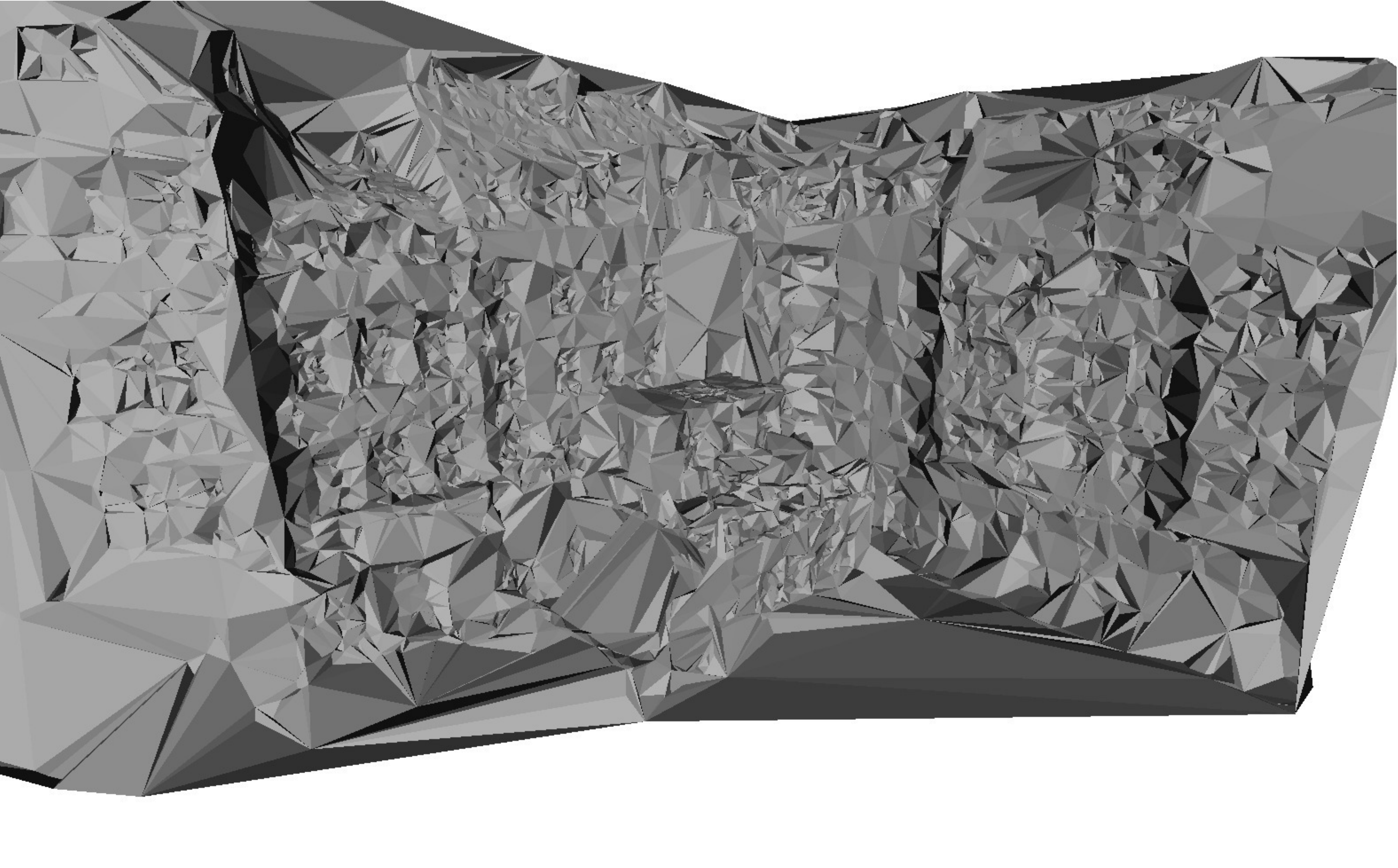}&
\includegraphics[width=0.14\textwidth]{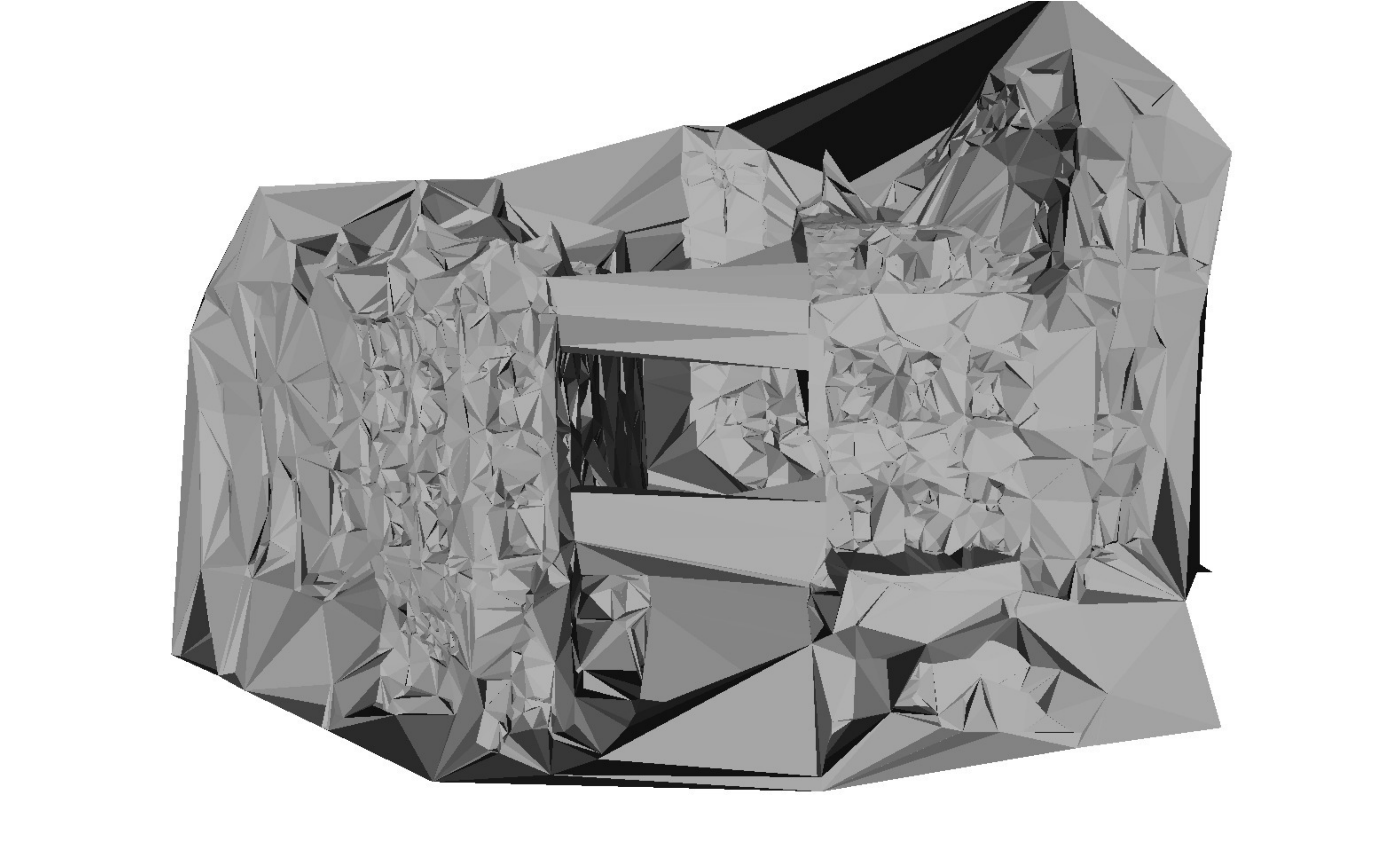}&
\includegraphics[width=0.14\textwidth]{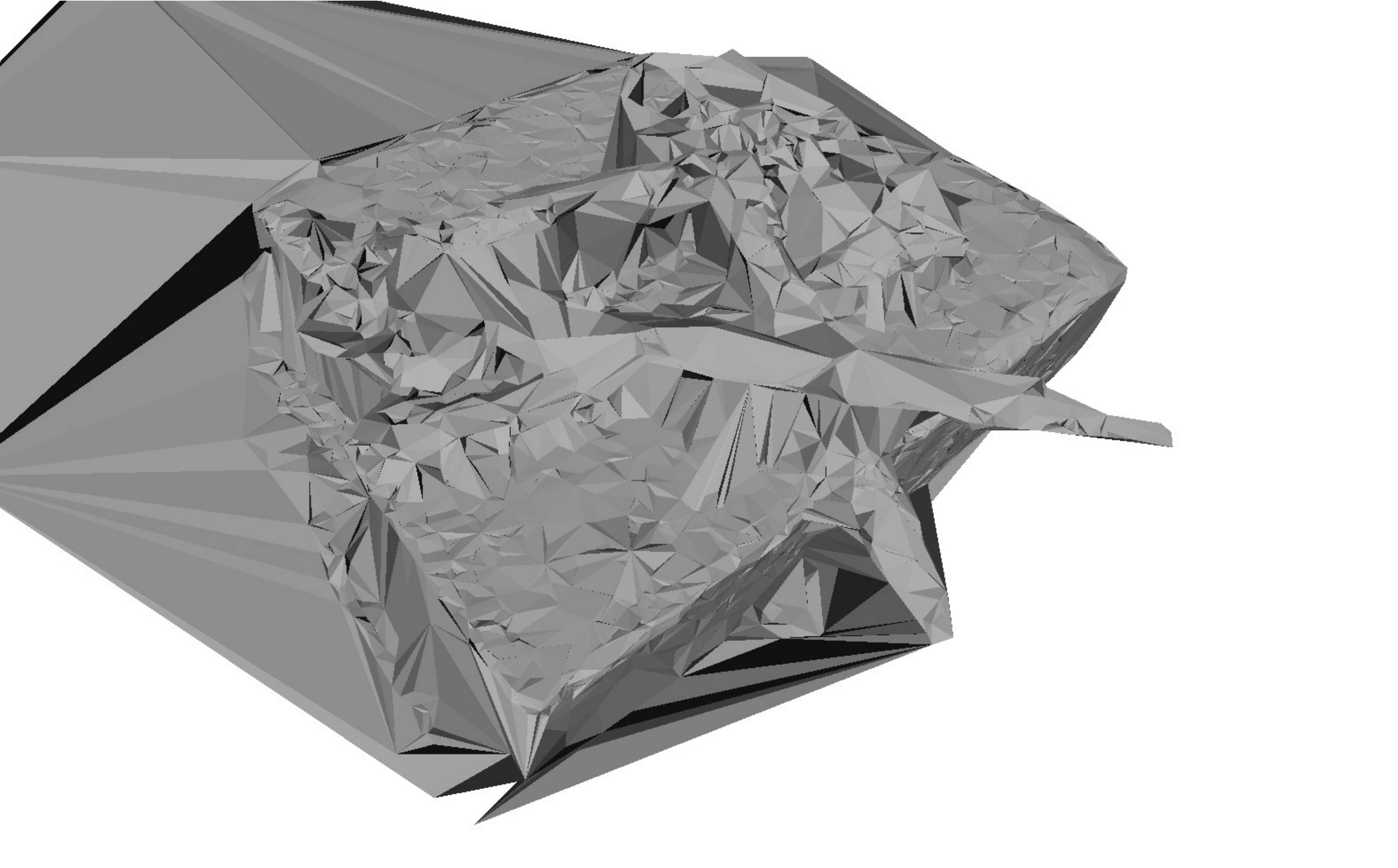}&
\includegraphics[width=0.14\textwidth]{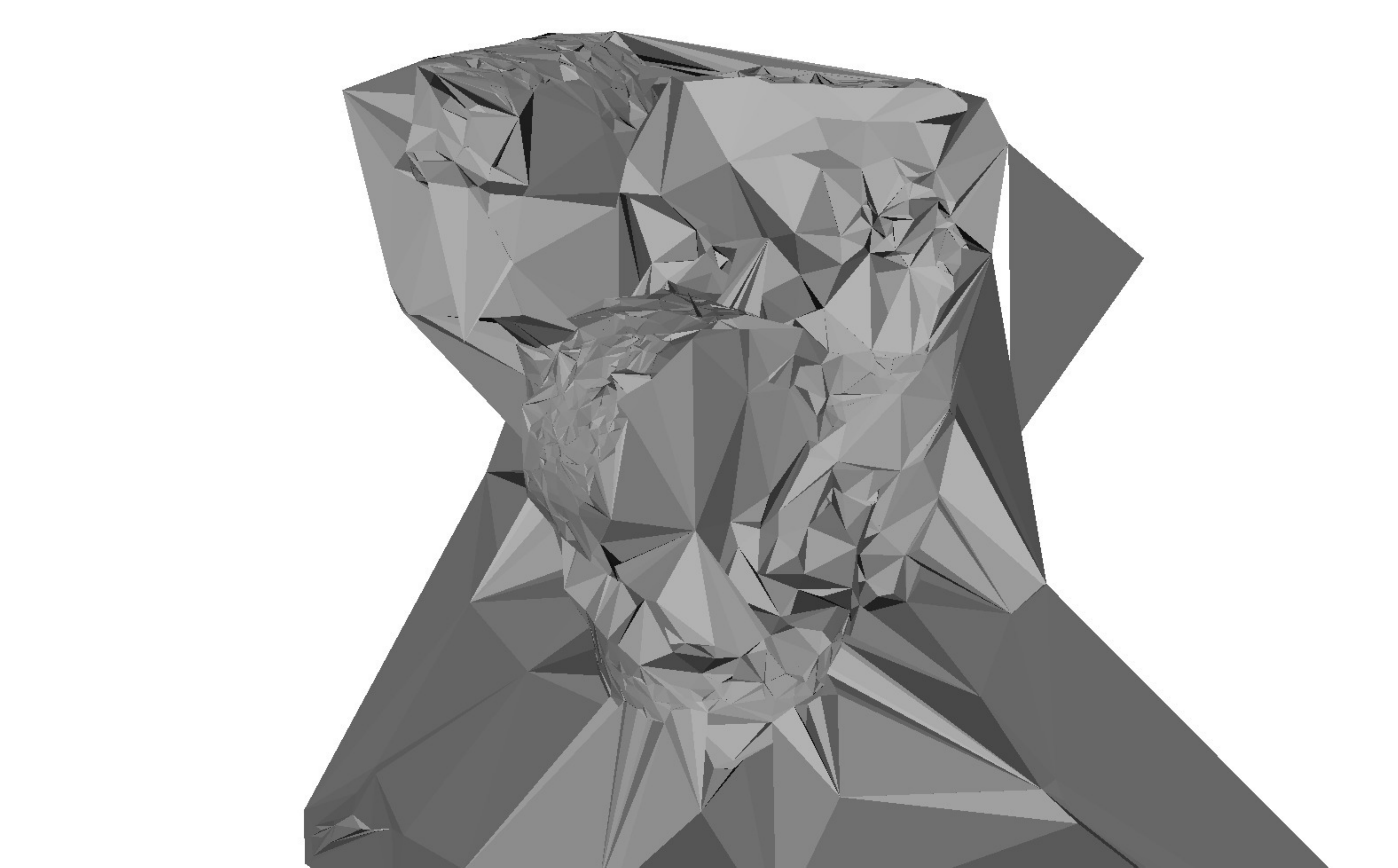}&
\includegraphics[width=0.14\textwidth]{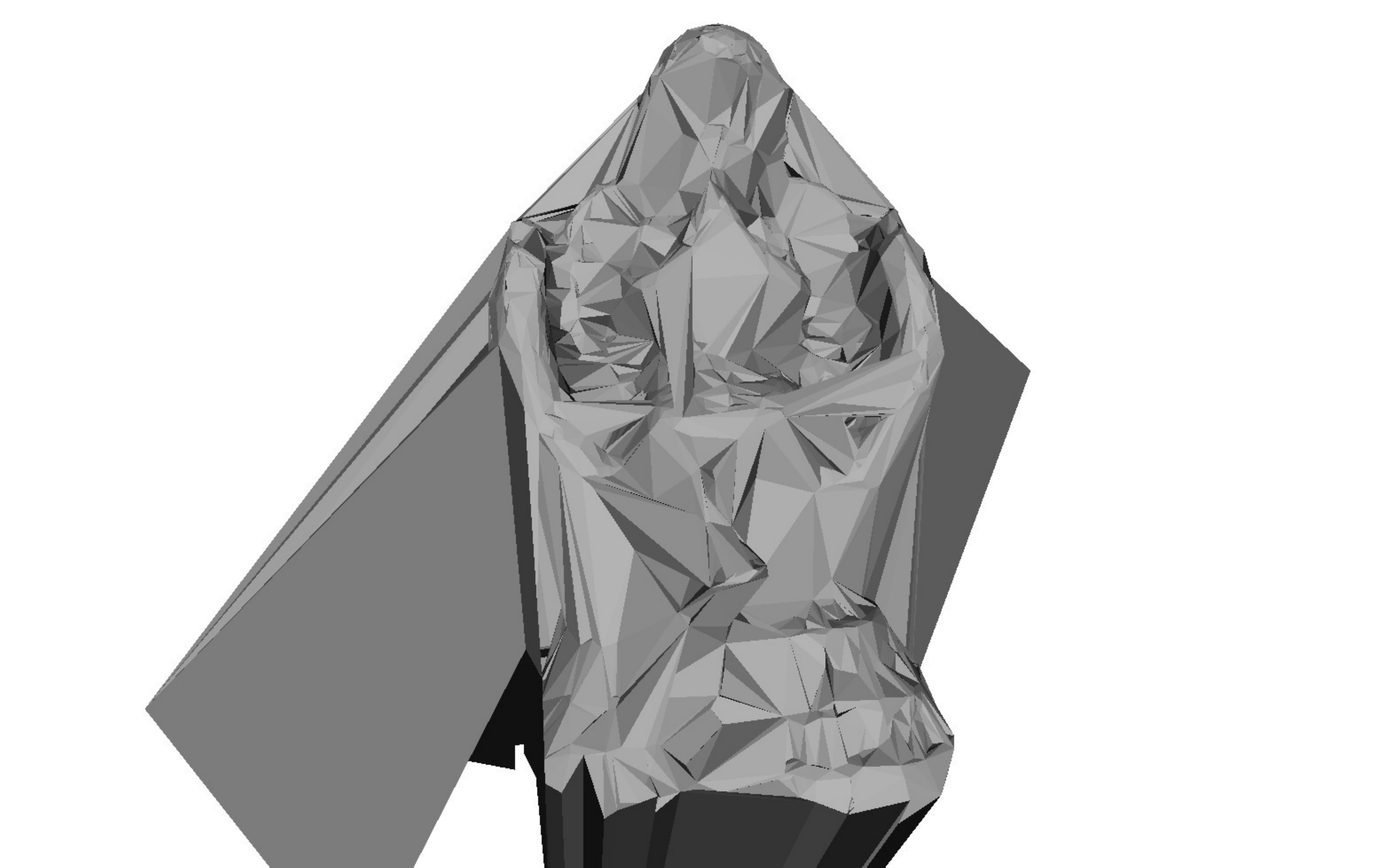}\\
\includegraphics[width=0.14\textwidth]{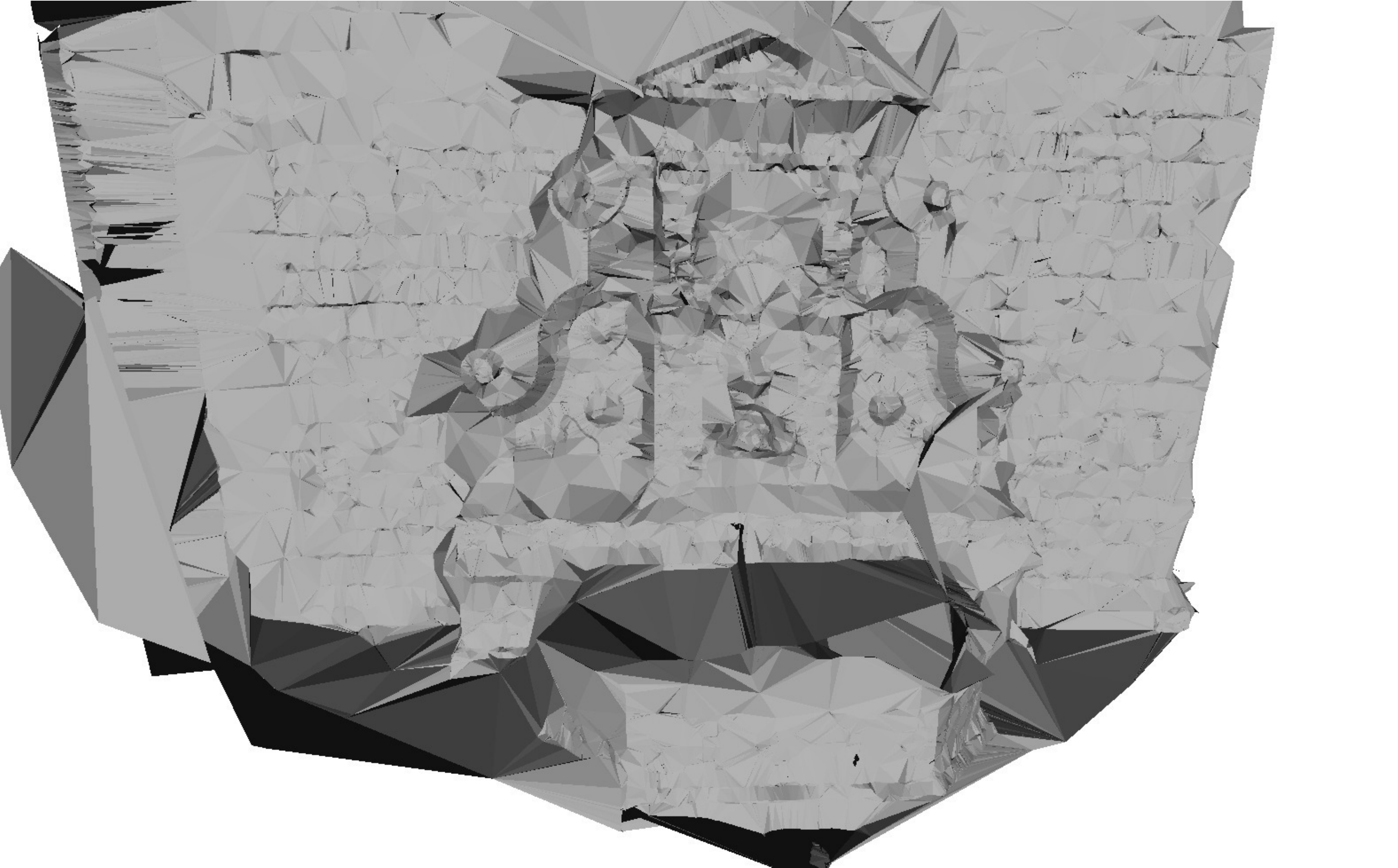}&
\includegraphics[width=0.14\textwidth]{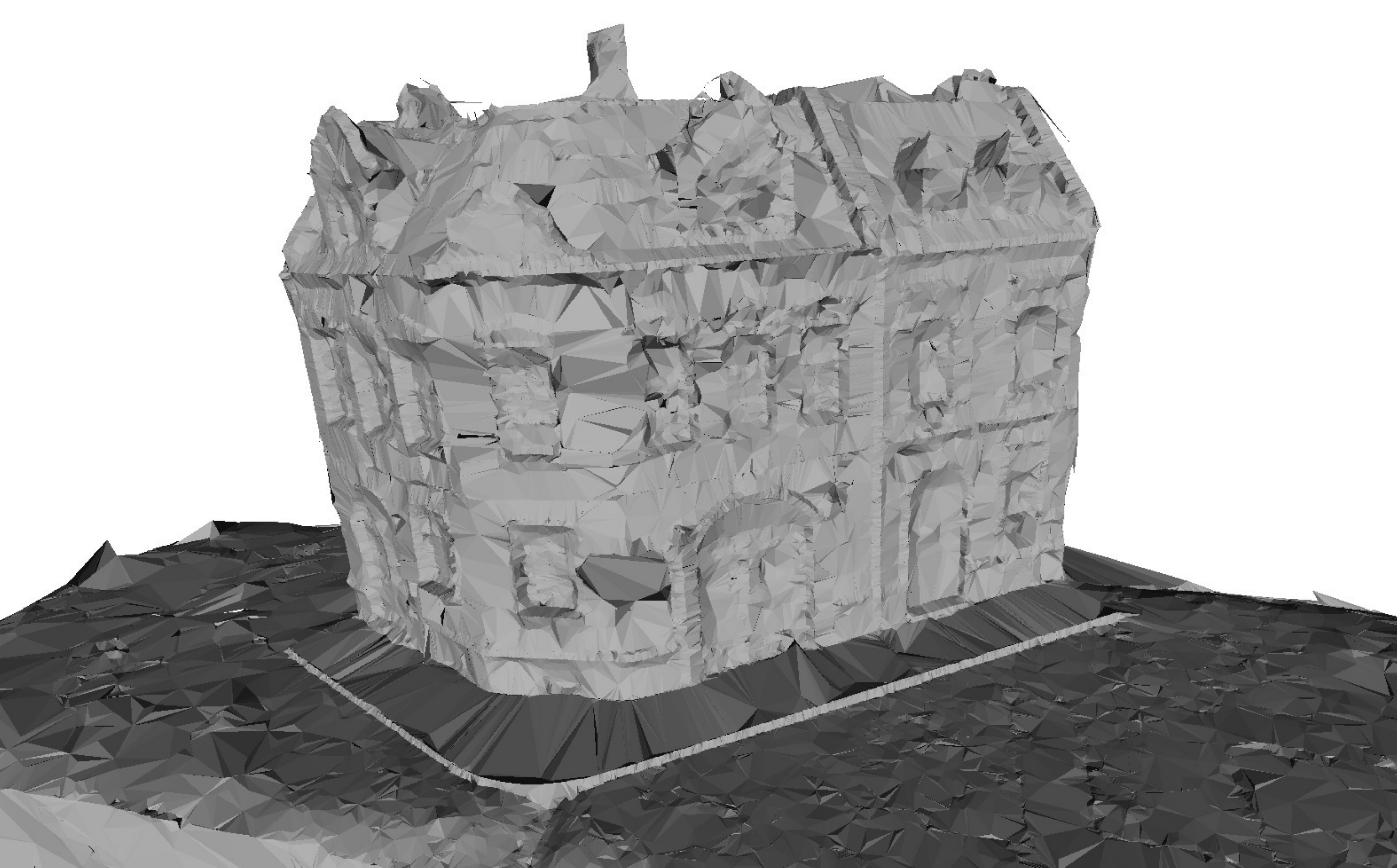}&
\includegraphics[width=0.14\textwidth]{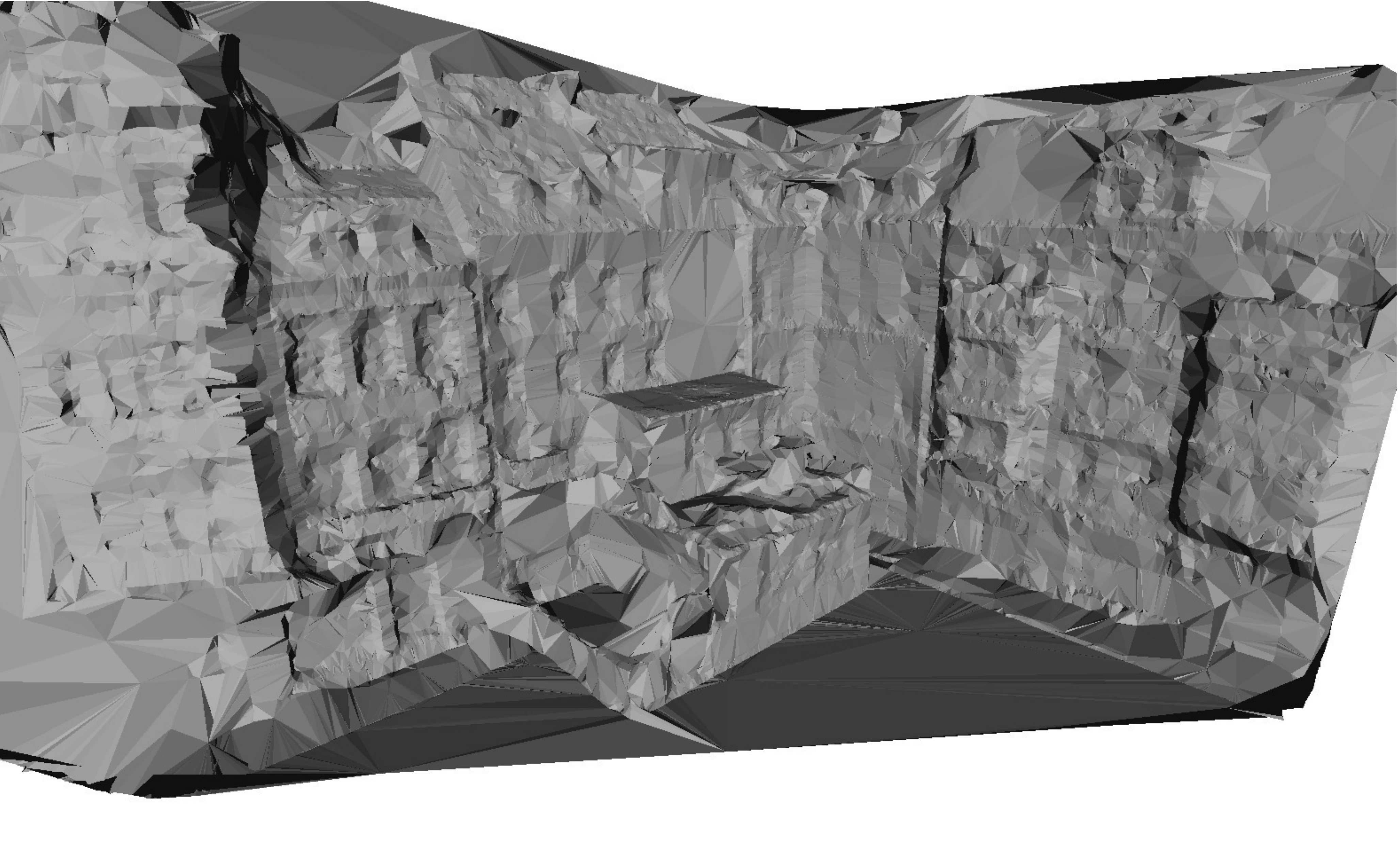}&
\includegraphics[width=0.14\textwidth]{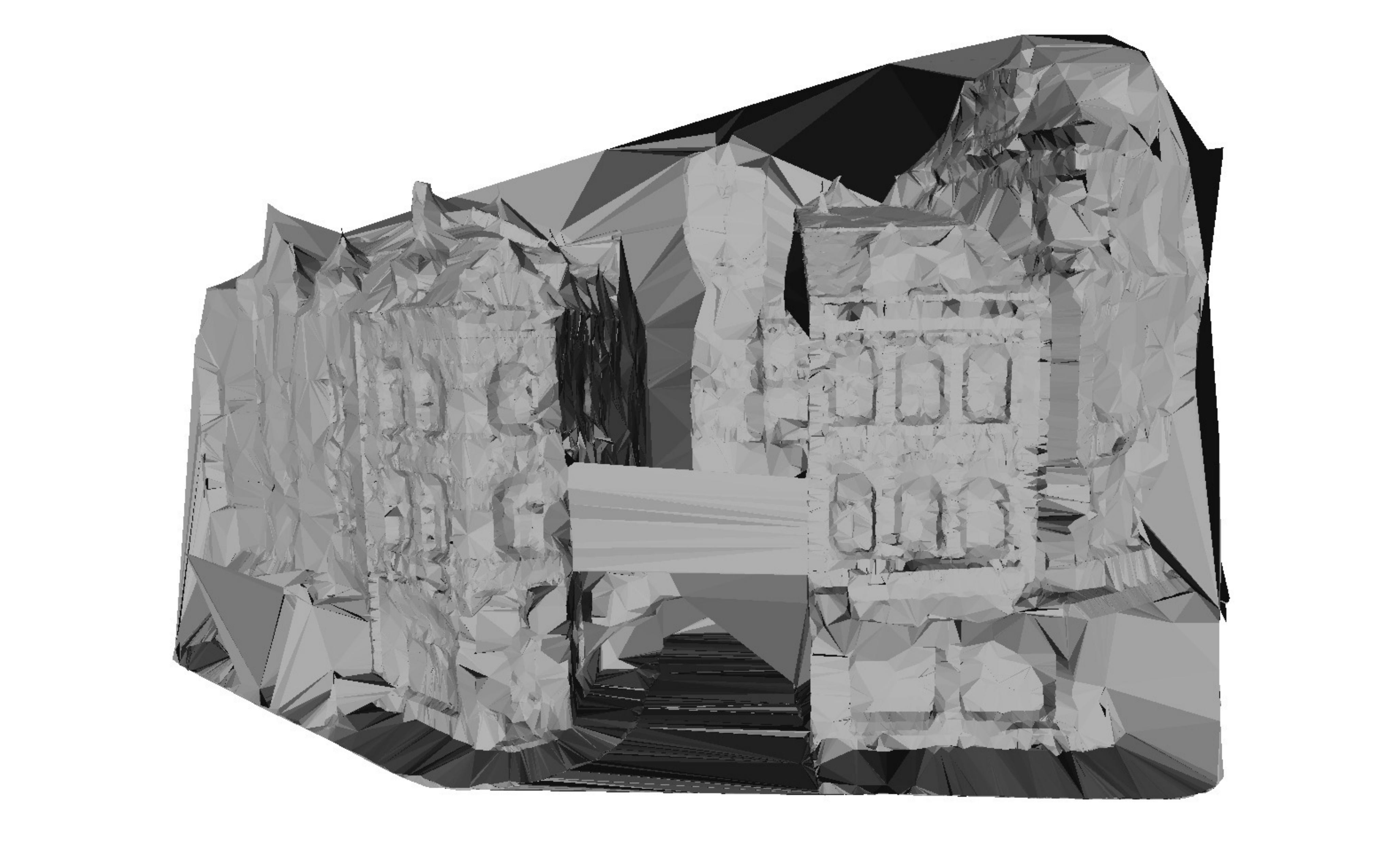}&
\includegraphics[width=0.14\textwidth]{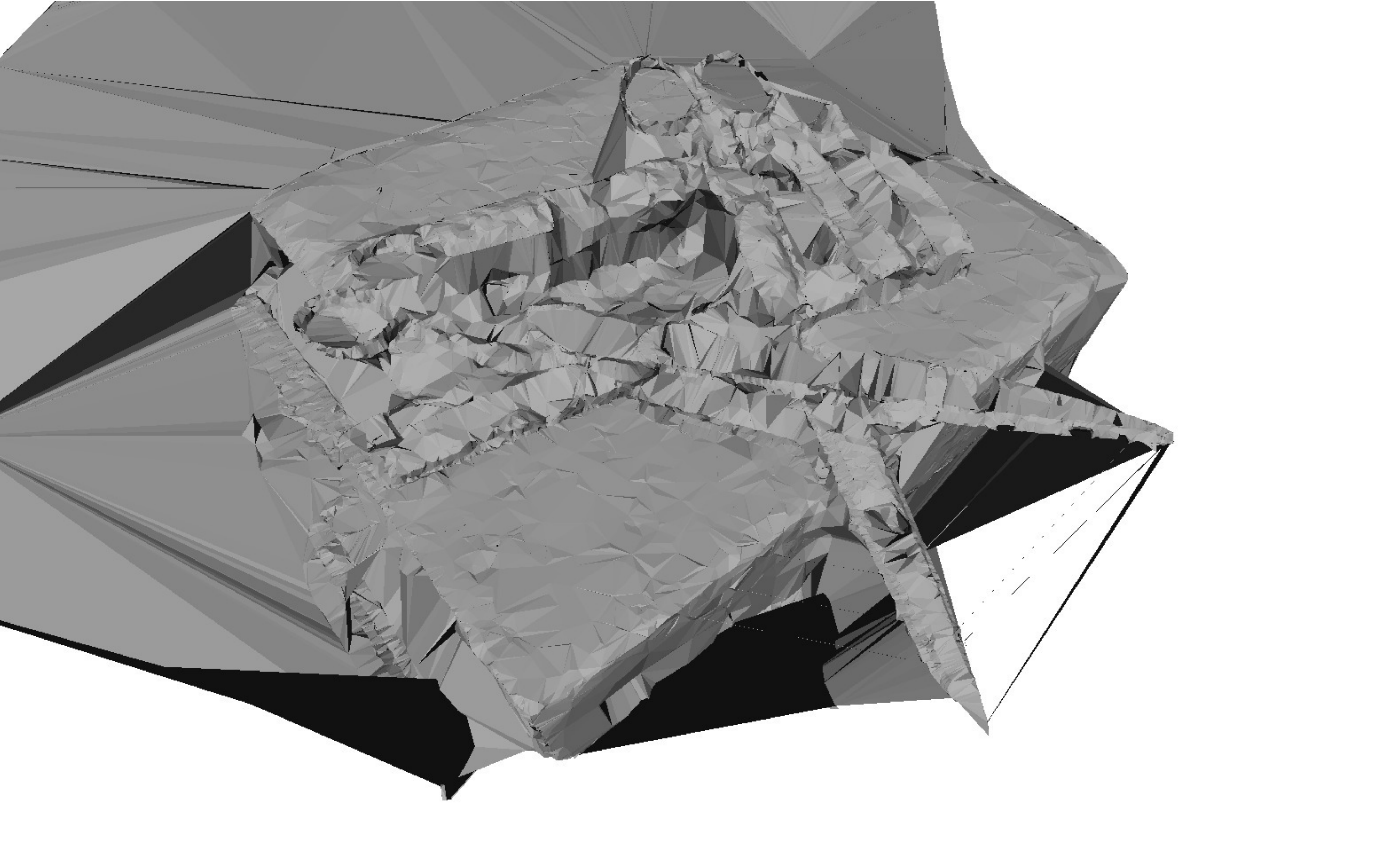}&
\includegraphics[width=0.14\textwidth]{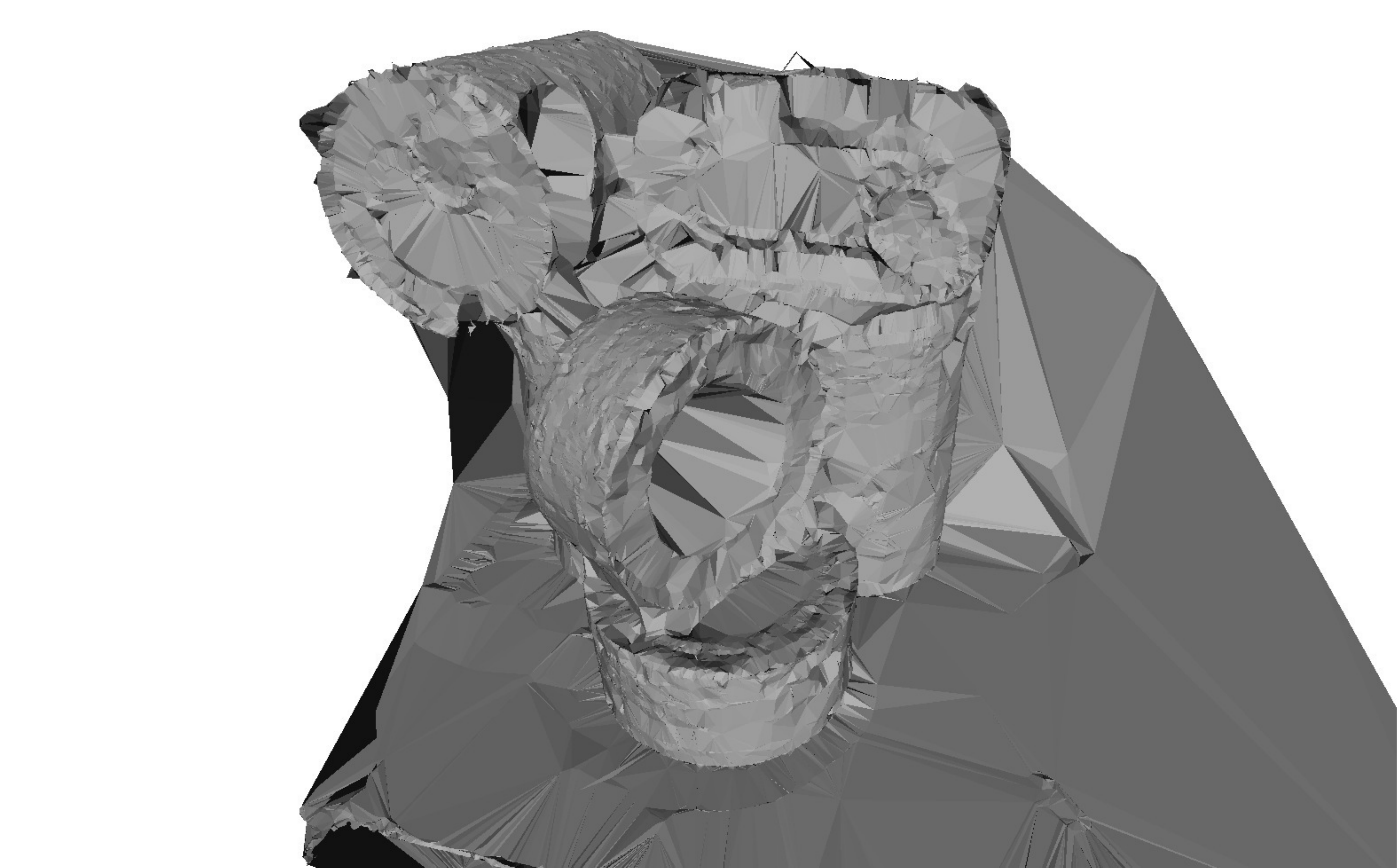}&
\includegraphics[width=0.14\textwidth]{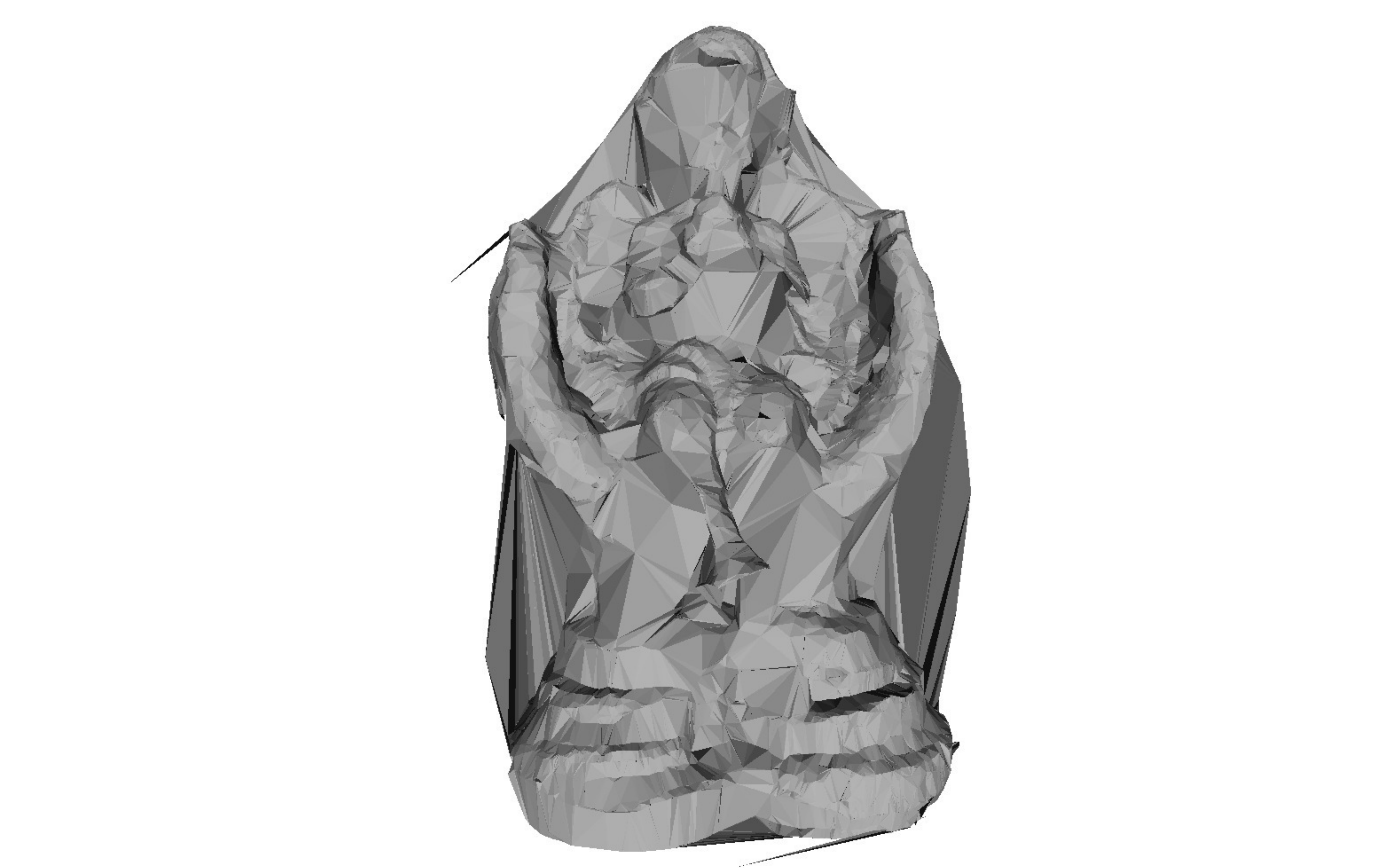}\\
\end{tabular}
\vspace{\belowdisplayskip}
\caption[blablabla]{
By row: ground truth, point cloud from OpenMVG, point cloud from our method, mesh from OpenMVG cloud, mesh from proposed cloud
}
\label{fig:resuCloud}
\label{fig:resuMesh}
\end{figure*}

Our algorithm bootstraps from the SfM point cloud generated by OpenMVG \cite{openMVG}, which provides very accurate points; therefore we compare the point cloud sampled from the 3D edges, against those estimated by OpenMVG by means of the CloudCompare  software \cite{cloudcompare}.
In Table \ref{tab:expRes} we list the Mean Absolute Errors (MAE), the Root Mean Squared Errors (RMSE) and the variance of the errors ($\sigma$).
As expected, we  significantly increase the number of reconstructed points (by one order of magnitude); despite our algorithm reconstructed full 3D edges, the accuracy of the reconstructed points remained close to the accuracy of the OpenMVG point cloud, which are easier to estimate, and in some cases this accuracy is even improved.
As Figure \ref{fig:resuCloud} shows, the proposed algorithm is able to recover structural elements that may remain completely undetected by the standard SfM process. 
For instance in DTU-006, we are able to reconstruct edges of any inclination,
recovering all the structural elements in the scenes; in the DTU-098 dataset, the high reflectivity of the metallic cans causes the SfM pipeline to fail in reconstructing a considerable portion of the surfaces, while the same areas are fully recovered by the proposed system.

\begin{table}[t]
\normalsize
\centering
  \caption{Comparison between meshes produced by \sysname and Line3D++ \cite{hofer2015line3d}.}
  \label{tab:expResLine}
\begin{tabular}{lrccccccc}
      &   & MAE & RMSE & $\sigma$ \\
\hline
\multirow{2}{*}{fountain-P11}&
Line3D++ & 101.8 & 272.5 & 252.7 \\
& \sysname & \textbf{64.59} & \textbf{159.4} & \textbf{145.7}  \\
\hline
\multirow{2}{*}{DTU-006}&
Line3D++ & \textbf{1.792} & \textbf{6.552} & \textbf{6.302}\\
& \sysname  & 2.805 & 8.497 & 8.021\\
\hline
\multirow{2}{*}{DTU-023}&
Line3D++ & 4.778 & 10.06 & 8.851\\
& \sysname & \textbf{3.207} & \textbf{7.898} & \textbf{7.218}\\
\hline
\multirow{2}{*}{DTU-028}&
Line3D++ & 21.10 & 60.19 & 56.38\\
& \sysname & \textbf{13.22} & \textbf{35.74} & \textbf{33.21}\\
\hline
\multirow{2}{*}{DTU-037}&
Line3D++ & \textbf{20.13} & \textbf{36.18} & \textbf{30.07}\\
& \sysname & 25.32 & 40.12 & 31.12\\
\hline
\multirow{2}{*}{DTU-098}&
Line3D++ & 17.69 & 47.22 & 43.79\\
& \sysname & \textbf{5.907} & \textbf{18.42} & \textbf{17.45}\\
\hline
\multirow{2}{*}{DTU-118}&
Line3D++ & 9.498 & \textbf{20.05} & \textbf{17.66}\\
& \sysname & \textbf{7.806} & 21.06 & 19.56\\
\hline
\bfseries average & & \multirow{2}{*}{\textbf{-15.6\%}} & \multirow{2}{*}{\textbf{-17.0\%}} & \multirow{2}{*}{\textbf{-17.2\%}}\\
\bfseries variation & & & & \\
\end{tabular}
\end{table}

Since the enhancement of 3D Delaunay-based mesh reconstructions is one of the most relevant reasons why we estimate 3D edges, we also compared the two 3D meshes reconstructed through the algorithm described in \cite{romanoni15b} from the OpenMVG points, from Line3D++ \cite{hofer2015line3d} and from the points sampled from the 3D edges. 
As suggested in \cite{strecha2008} we compare depth map generated by the reconstructed  and the ground truth meshes from the central camera of the sequence of each dataset.
Table \ref{tab:expRes}  shows that our algorithm extends to a multi-view stereo setting the hypothesis suggested in \cite{romanoni15b}: a Delaunay-based reconstruction significantly improves whenever we adopt 3D points belonging to 3D real world edges. 
Indeed, the meshes estimated from  3D edges points are considerably more accurate than the meshes computed with only SfM points (see Figure \ref{fig:resuMesh}).
Moreover Table \ref{tab:expResLine} shows that in general, in the context of 3D reconstruction our approach generates a point clouds that induce more accurate mesh with respect to the mesh reconstructed on the point cloud generated by Line3D++.
Execution times range from a minimum of 4 minutes (DTU-118), to a maximum of 30 (DTU-023), and average at of 13 minutes for the considered datasets.

\section{Conclusion and Future Works}
\label{sec:concl}
In this paper we proposed a novel method to estimate 3D edges and introduced \sysname: a system able to recover 3D edges in a scene observed in a set of views. 
The source code that implements the proposed system is also made available at \repolink.
While existing methods rely on video sequences and estimate only straight edges, our algorithm is able to recover straight and curved edges from an unordered set of images. 
We represent the image edges as edge-graphs and we match them according to epipolar and spatial constraints.
We also showed how Delaunay-based 3D reconstruction improves when built upon points sampled from reconstructed 3D edges.
As a future work we plan to integrate the 3D edges into the bundle adjustment process and to embed the recovered 3D edges into the reconstruction algorithm exploiting the Constrained 3D Delaunay triangulation.

\section*{Acknowledgements}
{\small
This work has been partially supported by the ``Cloud4Drones'' project founded by EIT Digital. }
{\small
\bibliographystyle{ieee}
\bibliography{biblioTotal}
}

\end{document}